\documentclass[journal, twoside]{IEEEtran}
\usepackage{amsfonts}
\usepackage{amsmath}
\usepackage{color}
\usepackage{enumitem}
\usepackage{cite}
\usepackage{soul}
\usepackage{balance}
\usepackage{booktabs}
\usepackage{ulem}
\usepackage[capitalise]{cleveref}
\ifCLASSINFOpdf
\usepackage[pdftex]{graphicx}
\else
\usepackage[dvips]{graphicx}
\fi
\begin{document}
	%
	\title{$\textup{F}^{\textup{3}}$A-GAN: Facial Flow for Face Animation with Generative Adversarial Networks}
	%
	%
	%
	
	\author{Xintian~Wu,
		Qihang~Zhang,
		Yiming~Wu,
		Huanyu~Wang,
		Songyuan~Li,
		Lingyun~Sun,
		and~Xi~Li*
		\thanks{X.~Wu, Q.~Zhang, Y.~Wu, H.~Wang, and S.~Li are with the College of Computer Science and Technology, Zhejiang University, Hangzhou 310027, China. (E-mail: \{hsintien, qh\_zhang, ymw, huanyuhello, leizungjyun\}@zju.edu.cn)}
		\thanks{L.~Sun is with the International Design Institute of Zhejiang University, Hangzhou 310013, China. (E-mail: sunly@zju.edu.cn)}
		\thanks{X.~Li (corresponding author) is with the College of Computer Science and Technology, Zhejiang University, Hangzhou 310027, China. (E-mail: xilizju@zju.edu.cn)}
	}
	\maketitle
	

	\begin{abstract}
		Formulated as a conditional generation problem, face animation aims at synthesizing continuous face images from a single source image driven by a set of conditional face motion. Previous works mainly model the face motion as conditions with 1D or 2D representation (\textit{e.g., action units, emotion codes, landmark}), which often leads to low-quality results in some complicated scenarios such as continuous generation and large-pose transformation. To tackle this problem, the conditions are supposed to meet two requirements, \textit{i.e.,} motion information preserving and geometric continuity. To this end, we propose a novel representation based on a 3D geometric flow, termed facial flow, to represent the natural motion of the human face at any pose. Compared with other previous conditions, the proposed facial flow well controls the continuous changes to the face. After that, in order to utilize the facial flow for face editing, we build a synthesis framework generating continuous images with conditional facial flows. To fully take advantage of the motion information of facial flows, a hierarchical conditional framework is designed to combine the extracted multi-scale appearance features from images and motion features from flows in a hierarchical manner. The framework then decodes multiple fused features back to images progressively. Experimental results demonstrate the effectiveness of our method compared to other state-of-the-art methods.
		
	\end{abstract}
	

	%
	\IEEEpeerreviewmaketitle

	\section{INTRODUCTION}
	%
	%
	%
	%
	
	\IEEEPARstart{A}{s} an important and challenging problem, face animation \cite{pumarola2020ganimation, tripathy2020icface, songsri2019face, thies2016face2face, siarohin2019first} aims at automatically synthesizing continuous face images from a single source image. It is formulated as a conditional generation problem that given a set of conditional variables describing face motion (\textit{e.g.}, expression or pose), the synthesis system is able to transform a source face image to the corresponding target images. For different types of conditions, this kind of condition-driven model has a wide range of applications in virtual actors generation (landmark \cite{wu2019landmark, ho2020interpolation}), and talking face generation (audio, text \cite{chen2019hierarchical, zhou2019talking, fried2019text, yu2019mining, su2019bridging})
	
	Considering the pipeline of such a conditional generative model for face animation, previous works mainly model the face motion in shape or expression using different representation such as 2D landmark \cite{qiao2018geometry, kossaifi2018gagan, song2018geometry, songsri2019face}, 1D action units \cite{pumarola2020ganimation, tripathy2020icface} or emotion codes \cite{ding2018exprgan, choi2018stargan, he2019attgan}. Despite their success in generating good results, some limitations still remain when adopting these conditions. Firstly, as mentioned above, conditions based on emotion codes only model the discrete expression while a natural face is supposed to lie in a continuous manifold. Secondly, these conditions may not benefit some special poses (\textit{e.g., a face profile}) because 1D or 2D representation leads to low-quality results including the representation accuracy and geometric structure in such situations. As a result, the synthesized images are still with artifacts or even irregular shapes. Therefore, in order to deliver a continuous sequence with more realistic effects, the conditions are supposed to meet two factors. One is the motion information preserving, \textit{i.e.,} recording the realistic face motion, the other is the geometric continuity, \textit{i.e.,} morphing the face continuously. To achieve these two goals, an intuitive idea is to directly perform 3D face modeling \cite{cao20193d} due to the fact that human faces change continuously in 3D space. In addition, modeling the geometry is also very important to preserve the shape consistency when morphing the face. 
	
	\begin{figure}[!t]
		\centering
		\includegraphics[width=\linewidth]{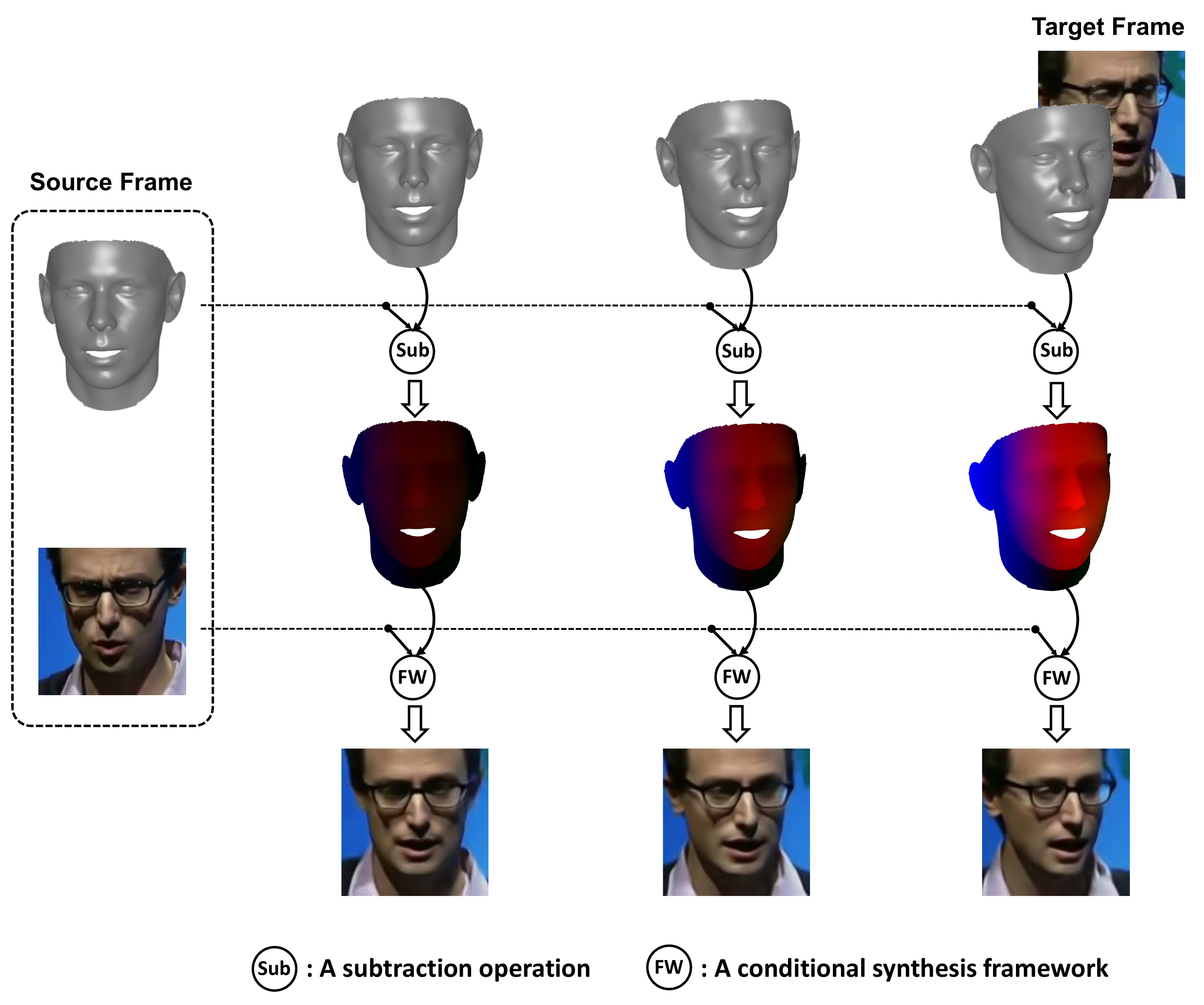}
		\caption{Diagram of our proposed method. Given the source and target image, we reconstruct their 3D face models and do interpolation to generate the intermediate face models. The facial flows are constructed through a subtraction operation on each pair of models. Then, the conditional synthesis framework automatically generates continuous face images driven by the continuous facial flows.} 
		\label{fig:first_pic}
	\end{figure}
	
\begin{figure}[!t]
		\centering
		\includegraphics[width=\linewidth]{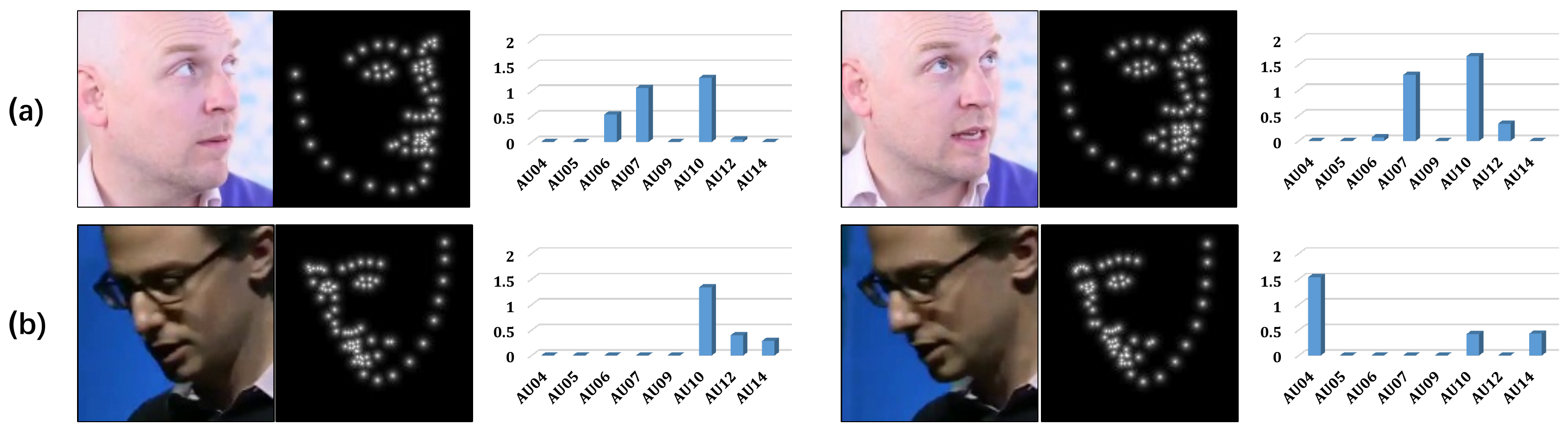}
		\caption{Illustration of using 2D landmarks and 1D action units (AUs) in profile images. The 2D landmarks are represented as a 2D landmarks heatmap with size of $W\times H\times 68$ on 68 landmark points, where $W$ and $H$ are the width and height of the images. The profile landmark heatmaps indicate that it is difficult to distinguish between different facial features (e.g., mouth, nose, and face outline). In (a), the results of AU values show that different expressions have similar AU activations in the profile pose. In (b), the results indicate that similar expressions corresponds to different AU activations.}
		\label{fig:second_pic}
	\end{figure}

	To this end, we propose a novel representation (see \cref{fig:first_pic}) based on a 3D geometric flow to represent the natural motion of the human face, namely \textit{facial flow}. It is constructed based on a 3D prior model 3DMM \cite{blanz1999morphable}, which is introduced in our work to reconstruct the 3D geometric information from a 2D face image through a linear transformation. The facial flow maps the face shape from a 2D plane to a 3D surface and is computed by storing the difference between two 3D face shapes into a 3-channel map. Compared with existing methods (see \cref{fig:second_pic}) in the literature, it is capable of better handling the two above requirements (\textit{i.e.,} motion information preserving, geometric continuity) in face animation. Firstly, by using 3D face reconstruction techniques, we model the flow in 3D space to preserve the face motion at any pose. Secondly, due to the linear continuity of the 3DMM parameter space, the facial flow well controls the continuous change to the face shape through parameters interpolation. 
	
	After constructing the facial flow, there exists another problem: \textit{how to use the facial flow for image editing}? To cope with this problem, we build a conditional GAN-based framework so that the source image can be guided by the facial flow in the feature space. To fully take advantage of the driven motion information from facial flow, we propose a hierarchical conditional framework, which extracts the features from image and flow branches separately. Then, it merges them at multiple scales in a hierarchical manner to guarantee both the low-level texture information and high-level semantic information in the fused results. All the multi-scale features are decoded back to images progressively. 
	
	We combine the two stages mentioned above as a workflow, termed $\textup{F}^{\textup{3}}$A-GAN, for face animation. It is capable of realizing various face animation tasks such as continuous face generation, expression reenactment, pose reenactment, \textit{etc}. Also, it is worth mentioning that our method realizes the continuous generation in a non-recursive way (see \cref{fig:first_pic}). We achieve the one-to-one generation of subsequent images from the first image to resist the error accumulation.
	
	
	The contributions of this work can be summarized as follows:
	\begin{itemize}[leftmargin=*]
		\item We propose a brand-new representation --- facial flow for face animation. It is a dense geometry-aware map containing 3D motion information for natural face movement.
		\item We propose a hierarchical manipulation strategy in a conditional GAN-based facial framework. It combines the multi-scale features of images and flows together hierarchically and decodes the features back to images progressively. 
		\item $\textup{F}^{\textup{3}}$A-GAN is capable of realizing various face animation tasks and experimental results show our scheme outperforms other synthesis methods in terms of both the visual quality and the diversity of generation.
	\end{itemize}


	%

	\section{RELATED WORK}
	
	\subsection{Conditional GANs in Face Manipulation}\label{II-A}
	
	Face manipulation \cite{tolosana2020deepfakes, ding2018exprgan, tang2019expression, choi2018stargan, huang2017beyond, wang2018high, wiles2018x2face, siarohin2019animating, zakharov2019few, ha2020marionette} is aimed at manipulating a single face image to a target face image driven by conditional motion information. Many early works \cite{de2010linear, wang2013aam, theobald2007real} use AAM \cite{cootes1998active} or 3DMM \cite{blanz1999morphable} to morph the face with artificial motion. These methods usually lack realism of the images because  synthesis systems edit the face in high-dimensional pixel space, which is difficult to achieve. Recently, GANs \cite{goodfellow2014generative, radford2015unsupervised, karras2019style, karras2020analyzing, zhu2020domain, isola2017image, hsu2019sigan, tang2020unified, ren2020deep} are widely used to generate fake images because of their powerful generating ability. Therefore, many conditional GAN-based methods \cite{mirza2014conditional, antipov2017face, pumarola2020ganimation, tripathy2020icface} have been widely studied in this task, and are grouped into several categories according to the types of conditions:  
	
	Firstly, Ding \textit{et al.} \cite{ding2018exprgan} and Tang \textit{et al.} \cite{tang2019expression} changed a face image to target expression conditioned on an emotion state. Choi \textit{et al.} \cite{choi2018stargan} proposed StarGAN to perform to-image translation for multiple domains conditioned on facial semantic attributes, generating different faces in hair, expression, gender, \textit{etc.} Secondly, in order to address the limitation of discrete representation used above, Pumarola \textit{et al.} \cite{pumarola2020ganimation} (GANimation) took advantage of AUs to generate anatomically-aware expression in a continuous domain. Based on it, Tripathy \textit{et al.} \cite{tripathy2020icface} added pose parameters in conditions for expression and pose reenactment. Thirdly, Qiao \textit{et al.} \cite{qiao2018geometry} and Kossaifi \textit{et al.} \cite{kossaifi2018gagan} added geometric constraints to the generative networks to maintain the face shape. Specifically, they extracted the sparse landmark points of the target face and transformed them into a spatial heatmap as a condition. These methods are capable of generating face images under arbitrary expression and pose but fail in some special cases such as face profile. Also, transferring the landmark from another person directly tends to output a deformed face. 
	
	
	\subsection{Continuous Generation}
	Different from face manipulation, the face animation task requires the animator not only to be able to generate faces of arbitrary expression but also to ensure continuous generation. In the following, we mainly depict several methods about the latter, which can be roughly divided into two categories:
	
	Firstly, continuous generation can be achieved through a conditional generative network with a set of continuous parameters as conditions. Among them, action units (AUs) are widely used to model the anatomical face movements of a human expression. They are anatomically related to the contractions of specific facial muscles. Many existing facial expression synthesis works \cite{pumarola2020ganimation, tripathy2020icface, zhou2017photorealistic} generate the new faces of target expression through GANs with the input concatenation of source face images and target AUs. 
	
	Secondly, recurrent neural networks (RNN) are incorporated into generative networks for video generation because they process the sequence data effectively. Tulyakov \textit{et al.} \cite{tulyakov2018mocogan} proposed mocogan to generate a video clip by sequentially generating video frames from decoupled content vectors and motion vectors. Songsri-in \textit{et al.} \cite{songsri2019face} (MotionGAN) proposed an LSTM-based generator and generated a face video from a single face image with an LSTM block element-wisely added to each frame generation. However, they are still not conducive to large-scale face transformation. This is because RNNs are weak in modeling long-term dependency. LSTM-based methods do not alleviate this problem well.
	
	
	
	\subsection{3D Face Reconstruction}
	
	The methods mentioned above learn an intrinsic representation of the shape and appearance of the faces in 2D space. Recently, 3D information is incorporated into deep generative techniques for fine-grained face manipulation. Geng \textit{et al}. \cite{geng20193d} took advantage of 3DMMs for face morphing and utilized GANs to integrate the face into the final output image. Ververas \textit{et al.} \cite{ververas2020slidergan} transformed an input face image into a new one according to continuous values of a statistical blendshape model of facial motion. It took advantage of the 3DMM parameters as conditions for conditional synthesis framework. Our method is similar to these but the difference is that we use the 3D model as prior guidance for the generative networks instead of fitting a large 3DMM parameter space based on a large amount of data. 
	
	To extract the 3D information from 2D images, 3D face reconstruction techniques are studied to reconstruct a face from an image into a 3D form (or mesh). Blanz \textit{et al}. \cite{blanz1999morphable} proposed a linear parametric 3DMM to model the shape and texture of 3D faces. It is learned from 3D scan of human heads with principal component analysis and expressed as a point cloud. Due to its linearity, a reconstructed model can be morphed through controlling the parameters as well.
	
	To fit the 3D model from an image, conventional methods \cite{levine2009state, booth20183d} obtain the proper parameters by iteratively solving the optimization problem. This process is relatively inefficient, and not suitable for real-time 3D face reconstruction. Song \textit{et al}. \cite{song2012three} proposed a RBF network modeling the intrinsic relationships between 3D models and 2D images to replace the iterative operation. Recently, CNN-based models are used to regress the 3DMM parameters due to their powerful fitting ability. Yi \textit{et al}. \cite{yi2019mmface} trained a CNN supervised by the ground-truth 3D annotations (\textit{e.g.}, parameters, 3D points), which are expensive to collect. Tu \textit{et al}. \cite{tu20203d} proposed a 2D-assisted self-supervised learning (2DASL) method that effectively estimates the 3D face without any 3D labels. Besides, instead of transforming the image to the parameter space, some works directly learn a complete 3D facial structure from image pixels. \cite{jackson2017large} performed a direct regression of a volumetric representation of the 3D face shape while \cite{feng2018joint} proposed a UV position map to record the position information of the 3D face.

	\section{APPROACH}
	
	In this section, we introduce the proposed $\textup{F}^{\textup{3}}$A-GAN to realize photo-realistic animation. Our model can be learned through a two-stage training scheme. In \cref{III-A}, we explain how to estimate the facial flow between any paired images with a CNN-based facial flow constructor. In \cref{III-B}, we introduce a hierarchical conditional framework to manipulate a source face image with given facial flows. In \cref{III-C}, we describe the training settings of these two stages.
	
	\begin{figure}[!t]
		\centering
		\includegraphics[width=\linewidth]{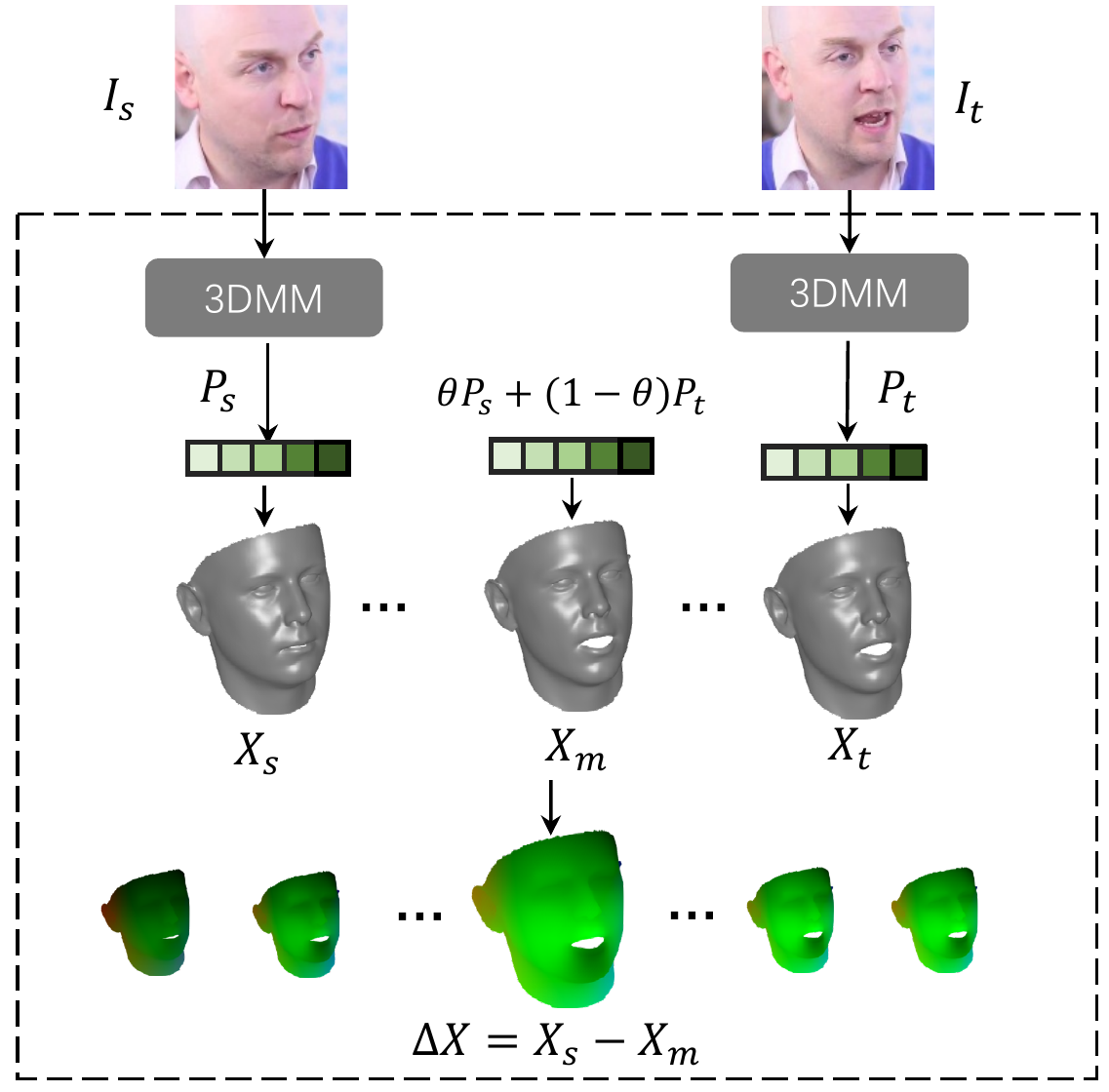}
		\caption{Illustration of the continuous facial flow generation. We reconstruct the source and target 3D face model $\{X_s, X_t\}$ from the corresponding input paired images $\{I_s, I_t\}$. Then, intermediate models are obtained by linear interpolation of 3DMM parameters.}
		\label{fig:stage1}
	\end{figure}
	
	\begin{figure*}[htbp]
		\centering
		\includegraphics[width=\linewidth]{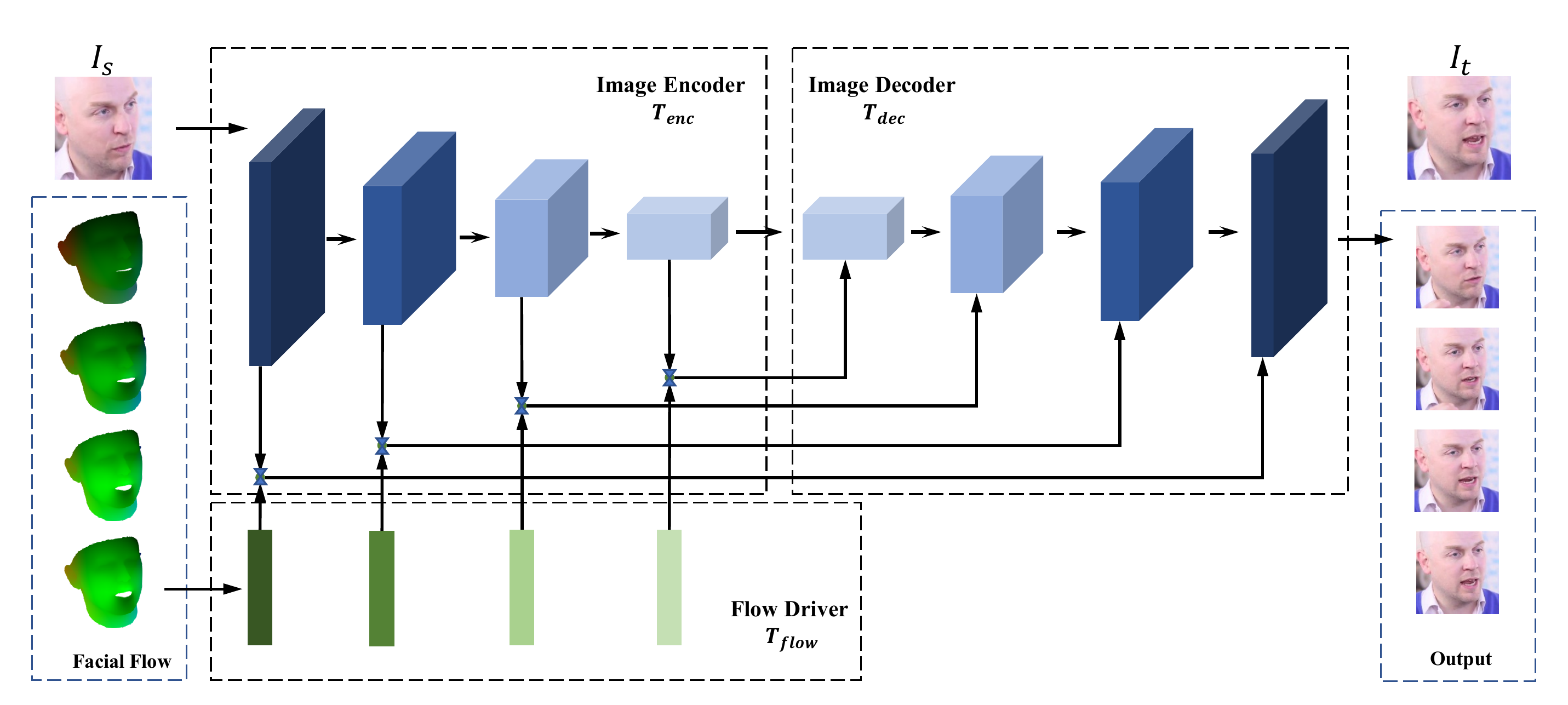}
		\caption{Diagram of our hierarchical conditional framework, which is a two-stream network. In the image stream, $T_{enc}$ and $T_{dec}$ are combined as an auto-encoder to extract the appearance features of the input image. In the flow stream, $T_{flow}$ aims to extract the geometric motion information from facial flows for driving manipulation.}
		\label{fig:framework}
	\end{figure*} 
	
	\subsection{Facial Flow Constructor}\label{III-A}
	
	We propose a facial flow constructor to estimate the intermediate continuous facial flows between a source $I_s$ and a target $I_t$ face image of the same identity for face animation. Since the facial flows are generated in a one-to-one mapping way, in the following, we first discuss how to estimate a single facial flow and then extend it to continuous facial flows.
	
	\normalem
	\subsubsection{Single facial flow estimation}
	To estimate a facial flow $f$ between $I_s$ and $I_t$, we first reconstruct the source $X_s$ and target $X_t$ face models of the corresponding face images through a learnable CNN, named 3DMM regressor. Then, we record the 3D motion between $X_t$ and $X_t$ in a 2D spatial map.
	
	In order to preserve the 3D geometric information, $f$ is represented as a 2D map recording the 3D motion in \{x, y, z\} axis in \{R, G, B\} channel, respectively.
	We take advantage of the 3DMM as a base model to fit a 3D face model from a 2D face image of any expression and pose. In specific, a morphable 3D face model can be formulated as:
	\begin{equation}
	X = s\ast R\ast (\overline{X} + A_{id}\alpha_{id} + A_{exp}\alpha_{exp}) + t,
	\label{con:reconstruct}
	\end{equation}
	where $X\in \mathbb{R}^{3N\times 1}$ is a 3D face model containing N vertices. Each vertex represents its coordinates ($x, y, z$) in three-dimensional space. $\overline{X}\in \mathbb{R}^{3N\times 1}$ is the mean shape of BFM \cite{paysan20093d} template model. $A^{id}\in \mathbb{R}^{3N\times K}$ is the identity base, usually extracted from the first $K$ principal components of facial scans with neutral expression and $\alpha_{id}$ is the corresponding identity parameter. $A_{exp}\in \mathbb{R}^{3N\times L}$ is the expression base, usually extracted from the first $L$ principal components of the discrepancy between neutral and expressive facial scans, and $\alpha_{exp}$ is the corresponding expression parameter. $R\in \mathbb{R}^{3\times 3}$ is the rotation matrix constructed from Euler angles $r\in \mathbb{R}^3$. $t\in \mathbb{R}^3$ is the translation vector and $s\in \mathbb{R}$ represents the scale scalar. The collection of all the model parameters is set as $P=\{ \alpha_{id}, \alpha_{exp}, r, t, s\}$. Through editing the coefficients in $P$, a face model of any expression and pose can be obtained. To align the 2D image with the 3D model, $X$ is often projected onto a 2D image plane through:
	\begin{equation}
	X_{2D} = \textbf{\emph{Pr}}\circ X,
	\end{equation}
	where, $\circ$ is the matrix multiplication, $\textbf{\emph{Pr}}$ is the orthographic projection matrix 
	$\left(               
	\begin{array}{ccc}   
	1 & 0 & 0\\ 
	0 & 1 & 0\\ 
	\end{array}
	\right)$. 
	
	Once we construct the 3D face models $\{X_s, X_t\}$ from the source and target images $\{I_s, I_t\}$, we can easily calculate the difference $X_s - X_t$ to represent the natural face motion in 3D space. However, all 3D points concatenated as a vector is not feasible for CNN processing and the flattened shape will lose the geometry. Therefore, we propose an image-like representation, which is also represented as a 2D map to store the information of this motion vector. In our settings, we use three flows $\{ f_x, f_y, f_z\}$ to save the motion along x,y,z axis, respectively. For each point $d^i_s = \{ x^i_s, y^i_s, z^i_s\} \in X_s$, and $d^i_t = \{ x^i_t, y^i_t, z^i_t\} \in X_t$, the motion along x, y, z axis will be $\Delta x^i_{st} = x^i_s - x^i_t$, $\Delta y^i_{st} = y^i_s - y^i_t$ and $\Delta z^i_{st} = z^i_s - z^i_t$. Then, we traverse each 3D point and project them onto the 2D plane according to the depth of $z$ dimension. As \cref{fig:first_pic}(Top) shows, the 3D motion $\{ \Delta x_{st}, \Delta y_{st}, \Delta z_{st}\}$ is stored in three channels $\{R, G, B\}$ of the facial flow map. Note that if two 3D points overlap on the x-y plane, we keep the one with larger depth value in z dimension, which is visible to the observers. 
	
	\subsubsection{Continuous facial flows generation}
	We convert the continuous natural movement of the face into the interpolation in 3DMM parameter space. Specifically, given $I_s$ and $I_t$, our 3DMM regressor is able to estimate their 3DMM parameters $P_s$ and $P_t$, respectively. As illustrated in \cref{fig:stage1}, the intermediate parameters can be linearly interpolated between the $P_s$ and $P_t$ well through \cref{con:interpolation} since the 3DMM parameter space is a linear space. 
	\begin{equation}
	P_m = \theta P_s + (1-\theta) P_t,
	\label{con:interpolation}
	\end{equation}
	where $P_m$ is the parameters of any frame in the middle, $\theta$ is the interpolation coefficient. Based on this, the intermediate face model $X_m$ can be reconstructed through \cref{con:reconstruct} and the 3D motion is calculated as $\{ \Delta x_{sm}, \Delta y_{sm}, \Delta z_{sm}\}$. Then, the intermediate facial flow is represented in the same way mentioned above. After estimating each facial flow individually, we apply them to the source image for continuous generation. This will be discussed in detail in the next subsection.
	
	\subsection{Hierarchical Conditional Framework}\label{III-B}
	
	We seek to estimate a mapping function $T: (I_s, f)\rightarrow I_t$ that transfers a source face image to a target face image with a conditional facial flow. In this section, we will describe the network architecture of our hierarchical conditional framework as follows:
	
	\subsubsection{Framework overview} Rethinking the pipeline of conditional GANs, we claim that the generator implicitly fuses the images and given conditions together to obtain new features. Since it is difficult for the network to learn such a mapping by directly concatenating images and driven flows as inputs, we try to extract their features separately and fuse them explicitly. As illustrated in \cref{fig:framework}, the framework is a two-stream network, which consists of an image encoder $T_{enc}$ and an image decoder $T_{dec}$ in the image stream, and a flow driver $T_{flow}$ in the flow stream. On the one hand, $T_{enc}$ and $T_{dec}$ aim to learn the appearance representation of an RGB image through self reconstruction. Given an input image, $T_{enc}$ extracts its multi-level features, and $T_{dec}$ tries to decode them back to itself. On the other hand, $T_{flow}$ aims to learn the geometric motion representation of the given facial flow. It utilizes the extracted motion features to reenact the appearance features of the source image through a manipulation module hierarchically, which will be described in the following. 
	
	\subsubsection{Hierarchical manipulation module} $T_{enc}$ and $T_{flow}$ extract features of different levels for the source image and conditional flow. In order to fully integrate the semantic information between different layers, we propose a hierarchical manipulation module embeddable in each layer and transitionable across layers. Specifically, in each layer, the appearance features $h_{enc}$ from $T_{enc}$ and motion features $h_{flow}$ from $T_{flow}$ of the same resolution are combined through the spatially-adaptive normalization (SPADE) \cite{park2019semantic}, which is able to maintain the spatial (\textit{i.e.,} geometric) information and widely used in style transfer. Then, inspired by \cite{karras2017progressive}, we transfer the learned generative ability from the deeper layer to shallower layer with a fade-in strategy. For example, given features of the $l$-th layer, the normalized output can be expressed as:
	\begin{equation}
	h_{out}^l=h_{dec}^l\cdot (\gamma(h_{flow}^l)\frac{h_{enc}^l-\mu(h_{enc}^l)}{\sigma(h_{enc}^l)}+\beta(h_{flow}^l)),
	\label{con:spade}
	\end{equation}
	where $\mu$ and $\sigma$ are the learnable parameters of the instance normalization, $\gamma$ is the new generated scale parameters and $\beta$ is the new bias. We combine the appearance features of the decoder together with element-wise multiplication. The output image is then formulated as 
	$\alpha \cdot Conv^{l-1}(DeConv(h_{dec}^l)) + (1-\alpha)\cdot Conv^{l-1}(Up(h_{dec}^l))$,
	where $Conv$, $DeConv$, $Up$ denote the convolution kernel, deconvolution kernel and linear interpolation upsampling, respectively.
	
	\begin{figure}[!t]
		\centering
		\includegraphics[width=\linewidth]{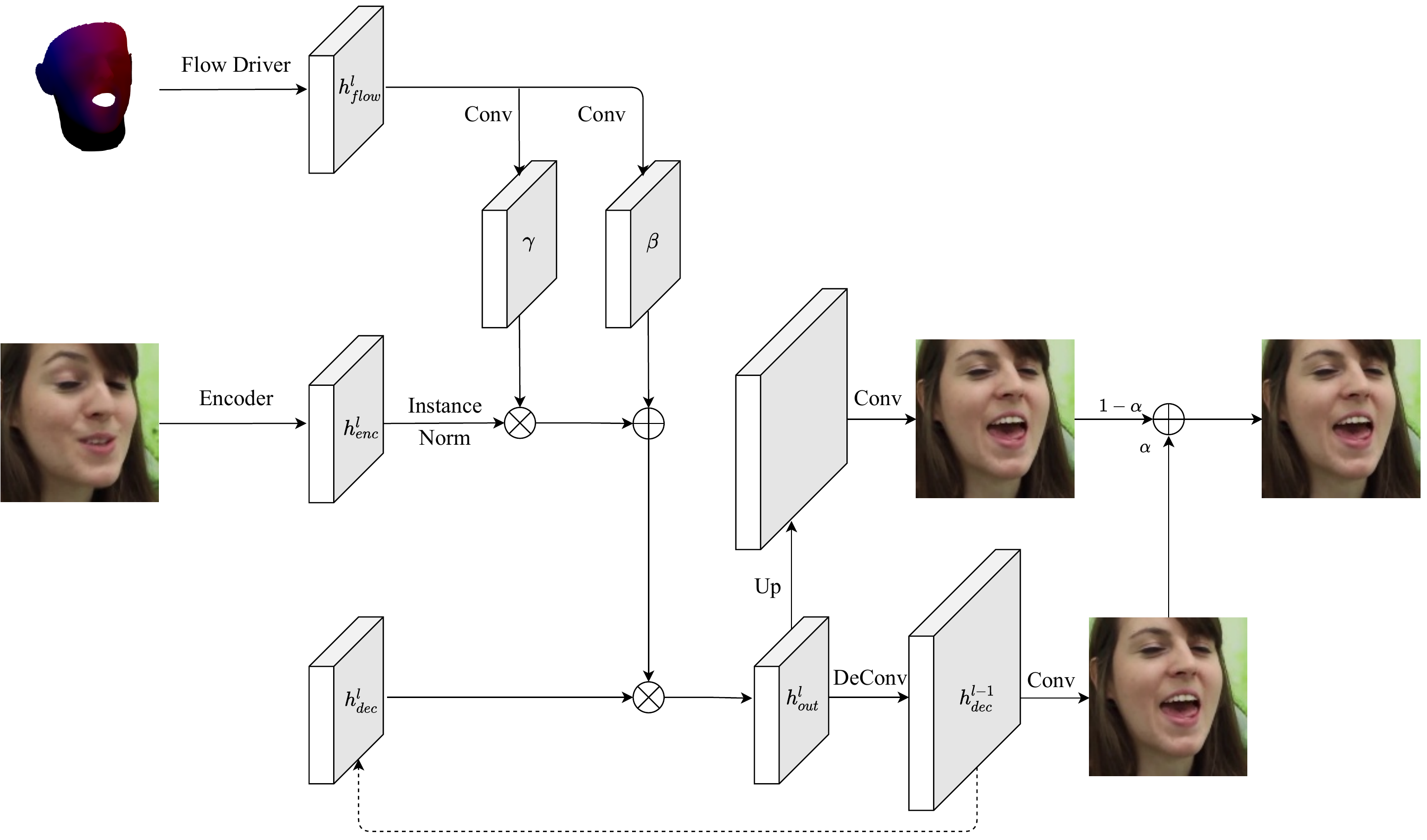}
		\caption{The architecture of our hierarchical manipulation module. The encoder features $h_{enc}^l$ of input images are normalized with the generated scale $\gamma$ and bias $\beta$ parameters from the features $h_{flow}^l$ extracted from facial flows. Then, the normalized features are multiplied by the decoder features $h_{dec}^l$. In order to enhance the performance of the decoder, we follow \mbox{\cite{karras2017progressive}} to generate the output images progressively with a fade-in strategy. In specific, we perform deconvolution and upsampling operations on $h_{out}^l$ at the same time to double the resolution.}
		\label{fig:manipulate_operator}
	\end{figure}
	
	\begin{figure*}[htbp]
		\centering
		\includegraphics[width=0.9\linewidth]{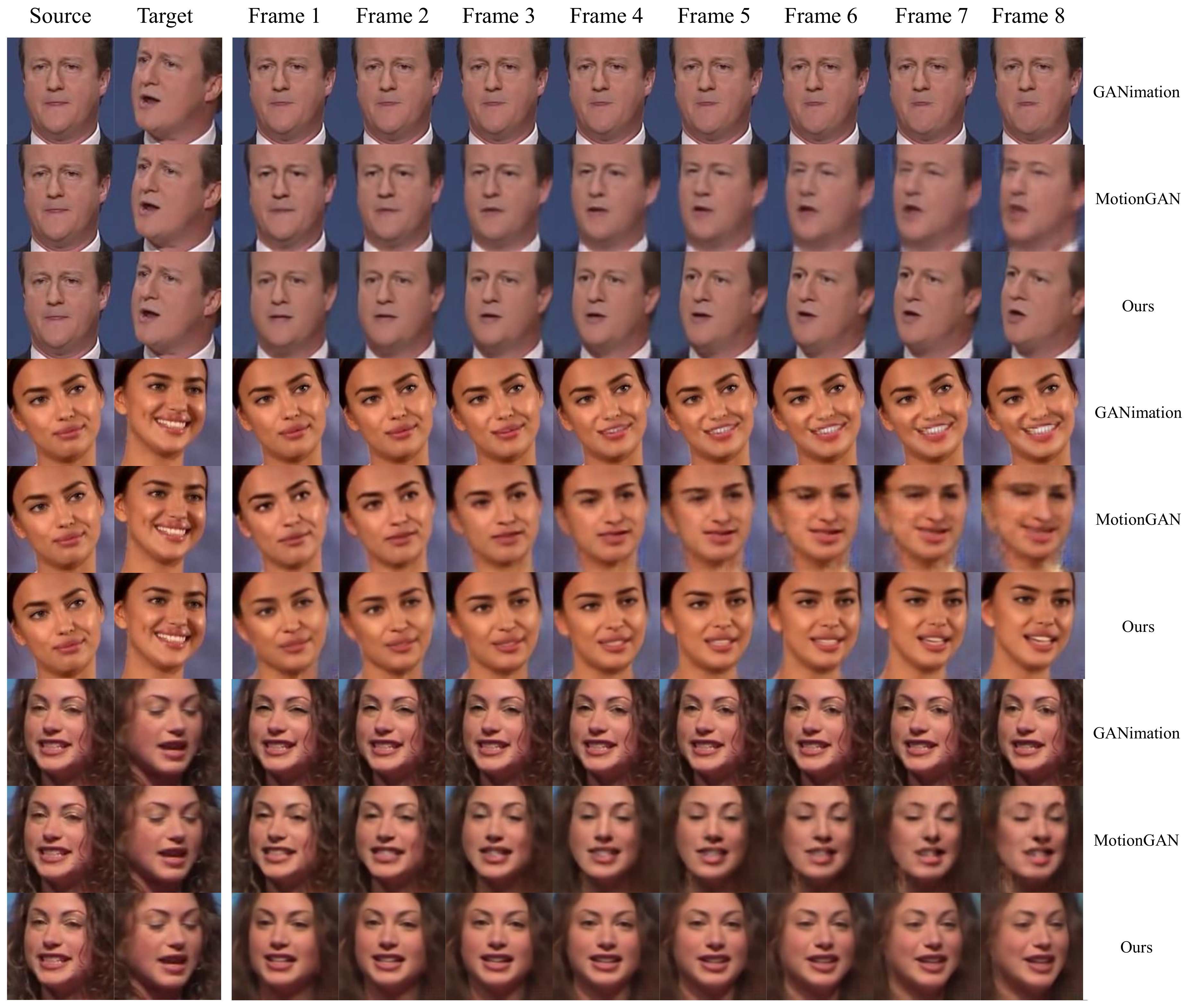}
		\caption{Qualitive comparison of continuous face generation by GANimation and MotionGAN. In these examples, we do interpolation on the corresponding conditions between two images $I_s$ (first column) and $I_t$ (second column) to generate the intermediate continuous images (from third column to the last column). }
		\label{fig:face_animation}
	\end{figure*}
	
	\begin{figure*}[!t]
		\centering
		\includegraphics[width=0.9\linewidth]{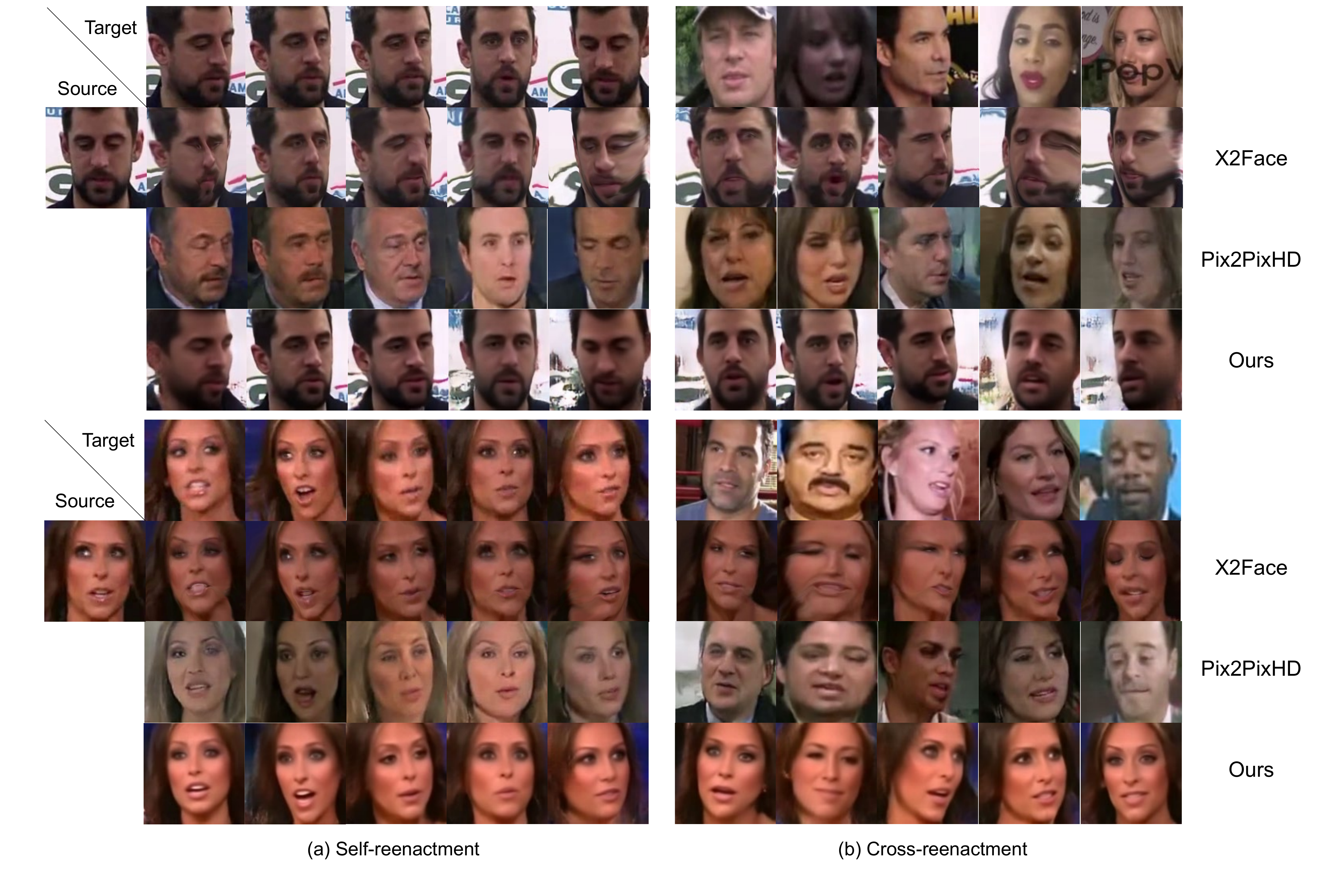}
		\caption{Qualitative results of self-reenactment and cross-reenactment on Voxceleb2 dataset. For self (a) / cross (b) identity reenactment, we randomly select few target images with the same / different identity as source image to drive the model generation. Each row indicates the generated images of the same method.}
		\label{fig:vox2_reenact}
	\end{figure*}

	\subsection{Training}\label{III-C}
	Our model is learned through a two-stage training scheme. In this section, we will describe the training settings of these two stages in the following. 
	
	\subsubsection{3DMM Regressor training} Our goal is to train a CNN-based 3DMM regressor to reconstruct the 3D face model of each frame in a video dataset. However, obtaining an accurate 3DMM regressor requires a large amount of training faces with 3D annotations, which are unaffordable in the video dataset. Training with a landmark consistency loss $L_{lm}$ \cite{tu20203d} does not perform well in continuous frames and easily suffers from a problem of mode collapse. 
	
	To remedy the problem analyzed above, we propose a new training scheme to deal with  sparse representation. Specifically, we introduce an image dataset with 3D annotations to warm up the network training. As for the task of 3D face reconstruction, we assume that conducting supervised learning with image datasets is relatively easy to achieve a good performance. Therefore, we first train our 3DMM regressor with a labeled image dataset and then transfer it to the unlabeled video dataset for self-supervised finetuning. Such a warm-up setting provides a better start point for the subsequent training.
	
	\textbf{Landmark loss $L_{lm}$}. The sparse landmark loss is utilized in most of the CNN-based methods and is defined as 
	\begin{equation}
	L_{lm} = \vert \vert w_{lm}\cdot (X_{2D}[:, \mathcal{L}] - \hat{M})\vert \vert_1,
	\label{con:landmark}
	\end{equation}
	where $X_{2D}$ is the projected face shape from an estimated 3D face model $X$, $\mathcal{L}$ is the indices of the landmark points in the 3D face model, $\hat{M}$ is the ground-truth landmark points (we also use superscript $\land$ to indicate \textit{ground truth} in the following), $w_{lm}$ is the weighted coefficients over different landmark points. We have increased the learning weights for the features that are difficult to learn such as eyes and mouth. 
	
	\textbf{Shape loss $L_{shp}$}. Directly learning various parameters at the same time will increase the training difficulty because the entire 3DMM parameter space is complicated. Therefore, we restrict the space by incorporating the ground-truth parameters. To this end, we optimize the identity parameter and expression parameter with $L_{shp}$, which is defined as  
	\begin{equation}
	L_{shp} = \vert \vert A_{id}\alpha_{id} + A_{exp}\alpha_{exp} - A_{id}\hat{\alpha_{id}} - A_{exp}\hat{\alpha_{exp}}\vert \vert_1.
	\end{equation} 
	
	\textbf{Transformation loss $L_{tr}$}. In addition, we regress the geometric transformation parameters $\{r, s, t\}$ directly. $L_{tr}$ is defined as
	\begin{equation}
	L_{tr} = \vert \vert r - \hat{r}\vert \vert_1 + \vert \vert s - \hat{s}\vert \vert_1 + \vert \vert t - \hat{t}\vert \vert_1
	\end{equation}
	
	In general, we train the 3DMM regressor with the following combined loss. In the warm-up training process, we conduct loss $L_{warm}$ for supervised pretraining and in the subsequent training process, we conduct loss $L_{sub}$ for self-supervised finetuning. Specifically, they are defined as 
	\begin{equation}
	L_{warm} = L_{shp} + \lambda_1 L_{lm} + \lambda_2 L_{tr} + \lambda_3 L_{reg},
	\end{equation}
	\begin{equation}
	L_{sub} = L_{lm} + \lambda_3 L_{reg},
	\end{equation}
	where $\lambda$'s are the weighting coefficients for different losses. $L_{reg}$ is a regularization loss that we add L1 regularization to regularize the identity and expression coefficients, which will prevent reconstructing a deformed face.
	
	\subsubsection{Conditional Framework training} As introduced in \cref{III-B}, we follow the pipeline of conditional GANs and propose a two-stream generator to manipulate a given face image with a target condition. Similar to other training settings of GANs, we adopt triplet data $\{I_s, I_t, f\}$ and a combined loss in an adversarial manner to train our conditional framework. Each loss will be described in detail as follows.
	
	\textbf{Weighted Pixel Loss $L_{wp}$}. Generally, an auto-encoder is optimized with a pixel-to-pixel loss for self-reconstruction. In order to enable the image stream to be expression-aware for face images, we adopt a weighted pixel loss to facilitate the network to pay more attention to the local facial features like eyes, nose, and mouth. We box out the facial feature according to landmark points and add higher weights on the mask of these areas. A weighted mask $w$ is then multiplied by the input image and the generated image. This loss is expressed as
	\begin{equation}
	L_{wp} = \vert \vert w\cdot (T(I_s, f) - \hat{I_t}\vert \vert_1,
	\end{equation}
	
	\textbf{Adversarial loss $L_{adv}$}. Adversarial loss is widely used in generation tasks. In our setting, a PatchGAN discriminator $D$ is used to measure the difference in distribution between two domains. Thus, the loss function can be expressed as
	\begin{equation}
	L_{adv} = D(\hat{I_t}) - D(T(I_s, f)),
	\end{equation}
	
	\textbf{Perceptual loss $L_{perc}$}. In order to keep the consistency of $\hat{I_t}$ and $T(I_s, f)$, we use the perceptual feature loss to shorten the discrepancy between these two in all layers. In addition, this loss can speed up the training process as well as reduce the blurring. Therefore, our perceptual loss is defined as 
	\begin{equation}
	L_{perc} = \vert \vert T_{enc}(T(I_s, f)) - T_{enc}(\hat{I_t})\vert \vert_1,
	\end{equation}
	
	\textbf{Total loss}. The complete loss function is obtained as a weighted combination of each loss defined above, which is defined as
	\begin{equation}
	L_{tot} = L_{adv} + \lambda_4 L_{wp} + \lambda_5 L_{perc}.
	\end{equation}

	\section{Experiments}
	
	This section provides a thorough evaluation for our framework. We first introduce the experimental setup in our method, including datasets, evaluation metrics and compared models. Then we compare our model against current competing techniques in the task of face animation and demonstrate our model's ability to deal with any-pose transformation and self / cross identity reenactment. Finally, we analyze our framework by conducting ablation studies on our proposed facial flow and network architectures. The implementation details and more experimental results are shown in the supplemental material.
	
	\begin{table*}[!t]
		\centering
		\caption{Quantitative evaluation of self-reenactment and cross-reenactment on voxceleb2 dataset. Upward/downward pointing arrows correspond to metrics that are better when the values are higher/lower.}
		\begin{tabular}{c|ccccc|ccc}
			\toprule
			& \multicolumn{5}{c|}{self-reenactment} & \multicolumn{3}{c}{cross-reenactment} \\
			\midrule
			Method & MSE $\downarrow$ & PSNR $\uparrow$ & SSIM $\uparrow$ & FID $\downarrow$ & CSIM $\uparrow$  & PRMSE $\downarrow$ & AUCON $\uparrow$ & CSIM $\uparrow$ \\
			\midrule
			X2Face [25] & 0.0116 & 21.0045 & \textbf{0.7641} & \textbf{22.3558} & 0.8988 & 5.0072 & 0.8527 & \textbf{0.8468} \\
			\midrule
			pix2pixHD [24] & 0.0550 & 13.4783 & 0.4904 & 35.9628 & 0.6996 & \textbf{2.2493} & \textbf{0.8744} & 0.6130 \\
			\midrule
			Ours  & \textbf{0.0100} & \textbf{21.7062} & 0.7426 & 28.8662 & \textbf{0.9082} & 3.6960 & 0.8445 & 0.8450 \\
			\bottomrule
		\end{tabular}%
		\label{tab:voxceleb2}%
	\end{table*}%
	
	\begin{table}[!t]
		\centering
		\caption{Quantitative evaluation of self-reenactment on 300VW dataset. Upward/downward pointing arrows correspond to metrics that are better when the values are higher/lower.}
		\begin{tabular}{ccccc}
			\toprule
			& MAE $\downarrow$ & MSE $\downarrow$ & CSIM $\uparrow$  & PRMSE $\downarrow$ \\
			\midrule
			GANimation & 0.1923 & 0.0966 & 0.8992 & 6.6313\\
			MotionGAN & 0.1815 & 0.0889 & 0.9211 & 6.1409 \\
			Ours-flow & \textbf{0.0165} & \textbf{0.0008} & \textbf{0.9821} & \textbf{0.7215} \\
			Ours-landmark & 0.0171 & 0.0010 & 0.9801 & 0.7619 \\
			Ours-aus & 0.0689  & 0.0139 & 0.8961 & 5.0906 \\
			\midrule
			$T_{os}$ & 0.0340 & 0.0035 & 0.9158 & 1.5368 \\
			$T_{np}$ & 0.0329 & 0.0036 & 0.9561 & 1.1984 \\
			$T_{sl}$ & 0.0277 & 0.0025 & 0.9502 & 1.1495 \\
			\bottomrule
		\end{tabular}%
		\label{tab:self-300VW}%
	\end{table}%
	
	\subsection{Experimental Setup}
	
	\subsubsection{Dataset}
	We introduce three datasets used in our experiments as follows. 
	
	\textbf{300VW}: We use 300VW dataset \cite{chrysos2015offline, shen2015first, tzimiropoulos2015project} for training and testing on face animation task. It is a face video dataset containing 114 videos with 104 people. All the frames in videos are cropped according to the head area with a size of $128\times 128$. Each image is labeled with the coordinates of 68 landmark points. We preprocessed the dataset by filtering out videos with poor image quality or a single expression, leaving 90 videos about 83 people in all. We randomly select 91000 paired images in the entire dataset for training and testing, respectively. In qualitative experiments, we randomly sample images in the whole dataset for manipulation.
	
	\textbf{300W-3D}: Since there are no 3D annotations in 300VW dataset, to improve the performance of constructing facial flow, we introduce 300W-3D dataset. It is an image dataset with 3D annotations of 300W \cite{sagonas2013300, sagonas2016300, sagonas2013semi} samples processed by \cite{zhu2016face}. Specifically, each image in 300W-3D dataset is annotated with 3DMM parameters including illumination, color, texture, shape, expression, and pose parameters. To align it with 300VW dataset, we detected the faces and box them out with a size of $128\times 128$. Then, we recalculated the translation and scale parameters to align the 3D face model with 2D face image. 
	
	\textbf{Voxceleb2}: In order to verify the effectiveness of our method, we also conduct several experiments on a larger dataset --- Voxceleb2 \mbox{\cite{chung2018voxceleb2}}. It is a publicly available video dataset containing more than 6000 celebrities extracted from YouTube, one million voice clips utterances. To speed up training, we only select 1000 identities for training and 119 identities for testing, each of which contains 10 videos for training. In all, we select 840100 paired images in the training set for training and 94466 paired images in the testing set for testing.
	
	\subsubsection{Metrics} 
	We conduct the experiments and evaluate our method on the following metrics. To evaluate the quality of the generated images, we use Mean Average Error (MAE), Mean Square Error (MSE) measuring the image similarity, Structure Similarity (SSIM) \mbox{\cite{wang2004image}} measuring the luminance, contrast and structure similarity, Fréchet-Inception Distance (FID) \cite{heusel2017gans} measuring the perceptual realism, Peak Signal to Noise Ratio (PSNR) measuring the signal energy error between generated images and ground-truth images.
	
	Under the scenario of cross-identity reenactment, we utilize other metrics to evaluate the model performance since ground-truth images are not provided. Cosine Similarity (SSIM) is used to measure the identity preservation of the generated images. It is calculated by the cosine distance between two embedded vectors generated by the pre-trained face recognition model (VGGFace) \mbox{\cite{parkhi2015deep}} on input images and generated images. Following \mbox{\cite{ha2020marionette}}, we compute PRMSE, the root mean square error of the head pose to measure the pose error, and AUCON, the ratio of identical action units to measure the expression error between the driving and generated images. Both the action units and head pose are estimated through OpenFace toolkit \cite{baltrusaitis2018openface}. 
	
	\subsubsection{Compared models}
	On the 300VW dataset we compare our method against two current conditional-GAN based methods with different conditions. GANimation \mbox{\cite{pumarola2020ganimation}}: it is capable of generating a wide range of continuous emotions and expressions for a given face image with action units as conditions. MotionGAN \mbox{\cite{songsri2019face}}: given a set of landmarks heatmaps of continuous frames as conditions, it takes advantage of LSTMs to map such a series of landmarks to images. For a fair comparison, we reproduce these two methods on 300VW datasets.
	
	On the Voxceleb2 dataset we compare our method against two other system. X2Face \mbox{\cite{wiles2018x2face}}: it consists of an embedding network for mapping the input face to an embedded face, and a driving network for generating a sampler to map the embedded face to the target face. We reused the released pre-trained model on Voxceleb1 for evaluation. Pix2PixHD \mbox{\cite{zhou2017photorealistic}}: it learns a mapping translating the input label map to the output face image with a coarse-to-fine generator. We follow \mbox{\cite{zakharov2019few}} and set the landmark heatmap as the input label map. Since no pre-trained models on Voxceleb dataset are released, we trained the model from scratch on Voxceleb2 dataset.
	
	
	\begin{table}[!t]
		\centering
		\caption{Metrics description.}
		\begin{tabular}{cc}
			\toprule
			Abbreviation & Description \\
			\midrule
			MAE/MSE & measuring the image similarity between \\
			PSNR  &  measuring the image signal energy error\\
			SSIM  &  measuring the structure similarity\\
			FID   &  measuring the perceptual realism\\
			CSIM  &  measuring the identity preservation\\
			PRMSE &  measuring the pose error\\
			AUCON &  measuring the action units error\\
			\bottomrule
		\end{tabular}%
		\label{tab:metric_descrip}%
	\end{table}%
	
	\subsection{Continuous Generation}
	
	We compare our model against other methods on both the 300VW and Voxceleb2 dataset for continuous generation. \mbox{\cref{fig:face_animation}} illustrates the results of 300VW dataset and more results on Voxceleb2 dataset can be seen in Section II in the supplemental material. Given any two face images $\{I_s, I_t\}$ of the same identity, we verify the difference of these methods by interpolating the intermediate sequence frames. Each row indicates the generated continuous frames between the given source face image and the target one. 
	
	It can be seen that our method outperforms the other two on the quality results because the proposed facial flow has dense geometric-aware characteristic and provides abundant information for generating faces of any expression and pose. Moreover, our method is able to generate continuous face images performing changes on both expressions and poses while maintaining the source identity relatively well. As for the compared methods, on the one hand, GANimation cannot output faces with pose changes since action units only model the expression action of the human face. Additionally, data-driven methods on action units recognition encountered a performance bottleneck and thus the values of action units extracted by some expressions are not accurate enough. This leads to a fail case as shown in the first person in \cref{fig:face_animation}. On the other hand, MotionGAN is capable of generating good results on the neighbor frames but fails in large-pose transformation. As illustrated in the last few columns of each second row, the face pose is preserved but details on face expression are missing. 
	
	\begin{figure}[!t]
		\centering
		\includegraphics[width=\linewidth]{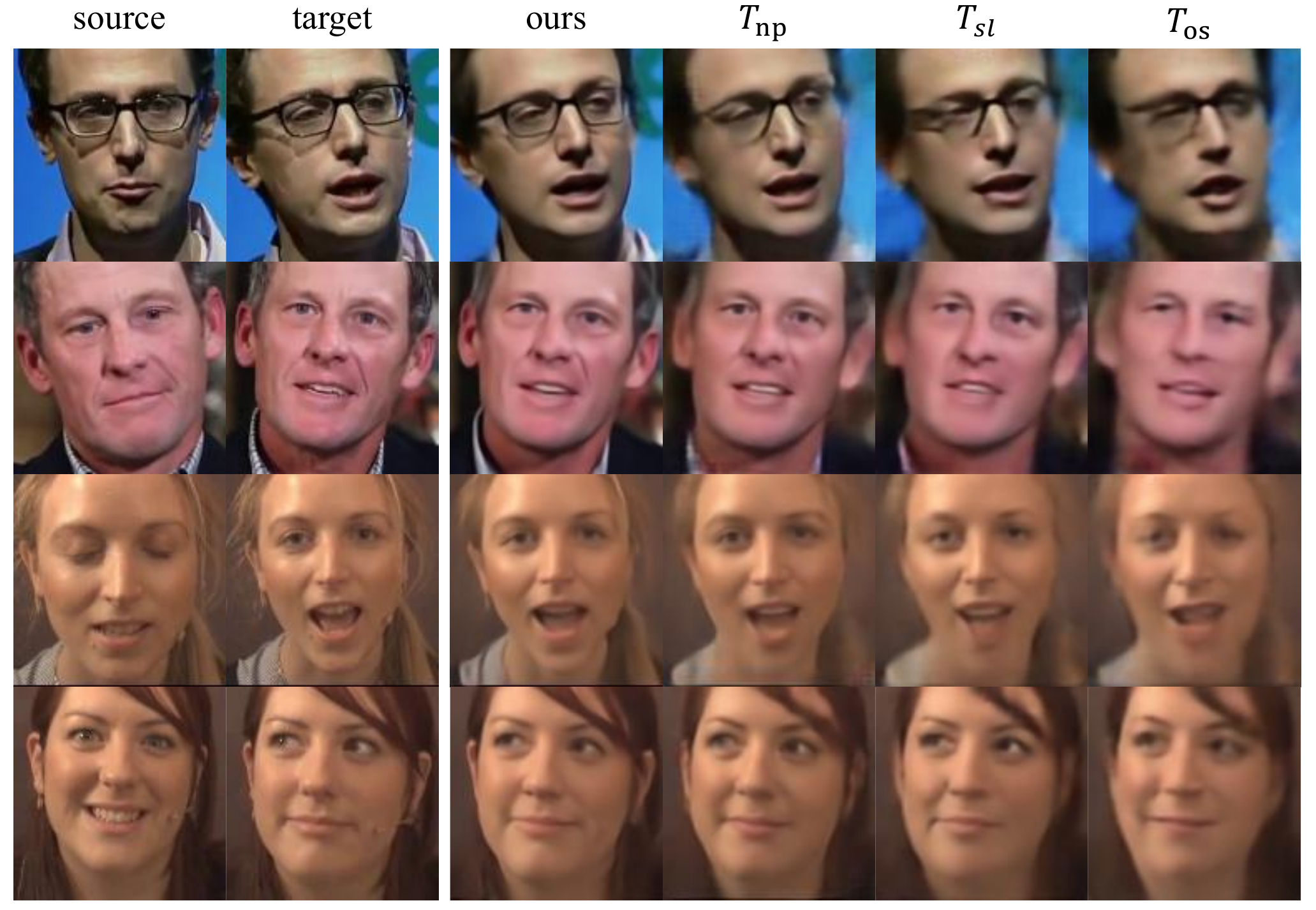}
		\caption{Visualization results of the manipulated output images based on different architectural variants. We do ablation study on the architectural variants of our proposed method. Here, $T_{np}$ represents the framework without progressive generation, $T_{sl}$ corresponds the framework only manipulates the image in single layer, and $T_{os}$ stands for the one-stream framework.}
		\label{fig:variants}
	\end{figure}
	
	\subsection{Self-Reenactment}
	In this subsection, we show the comparison results qualitatively and quantitatively under the self-reenactment setting \mbox{\textit{i.e.}}, reenacting the input face using the driving face of same identity. As shown in \mbox{\cref{fig:vox2_reenact}}-(a), our method synthesizes more accurate and realistic images. As for two compared methods, X2Face is on par with ours but fails to transform well in large pose (See the first and third image of the 1st row on the upper-left part). Additionally, Pix2PixHD is able to transform the pose and expression well but fails to keep the identity since it has poor generalization ability for unseen identities. This is due to that input of pix2pixHD is only the target sparse landmark heatmap without any appearance information of the source identity. More examples can be seen in Section II in supplemental material.

	Moreover, quantitative results on Voxceleb2 and 300VW dataset can be seen in \mbox{\cref{tab:voxceleb2}}-(left part) and \mbox{\cref{tab:self-300VW}}, respectively. In \mbox{\cref{tab:voxceleb2}}, we test the test set with five metrics, MSE, PSNR, SSIM, FID, CSIM. It can be seen that our method outperforms pix2pixHD on each metric and is on par with X2Face. Although X2Face is slightly better than us in FID and SSIM, it is prone to errors when performing pose transformation. It could be the reason that X2Face transforms pixels in image-level while ours drives the image with facial flow in feature-level. Our method also outperforms it in MSE, PSNR, CSIM in image quality estimation. In \mbox{\cref{tab:self-300VW}}, the quantitative results also indicate that our model generate high-quality images, and the high CSIM and low PRMSE values indicate that our model is able to keep the identity and pose relative well.

	
	\begin{figure}[!t]
		\centering
		\includegraphics[width=\linewidth]{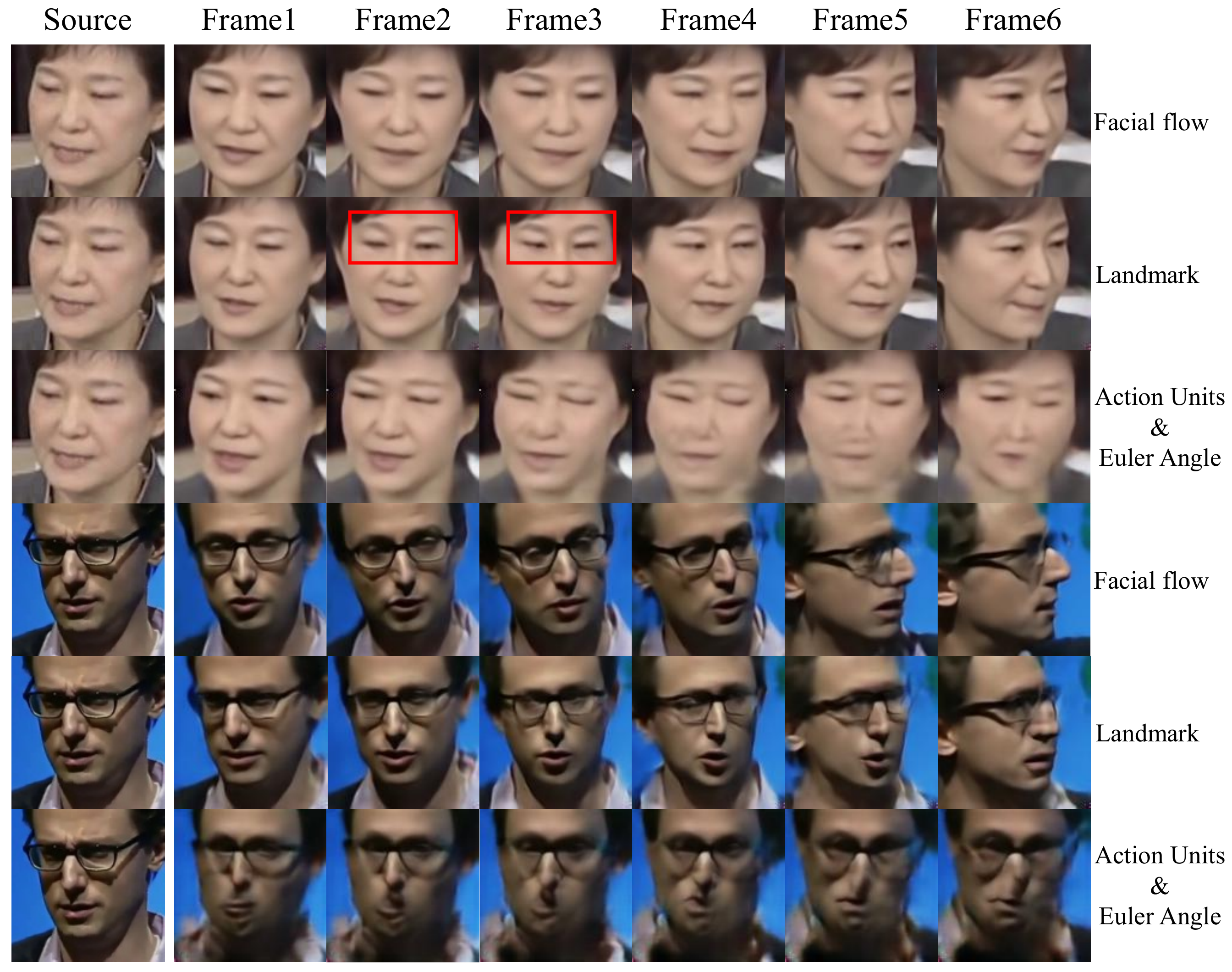}
		\caption{Comparison results of conditions study. In these examples, we also do interpolation on different conditions. For each source image, the first row is animated with facial flow as conditions, the second row is with landmark, and the third row is with action units. Images in the leftmost column are the ground-truth source images, while images demonstrated from the second column to the last column are synthesized by the hierarchical conditional framework.}
		\label{fig:conditions}
	\end{figure}

	\begin{figure*}[!t]
		\centering
		\includegraphics[width=0.95\linewidth]{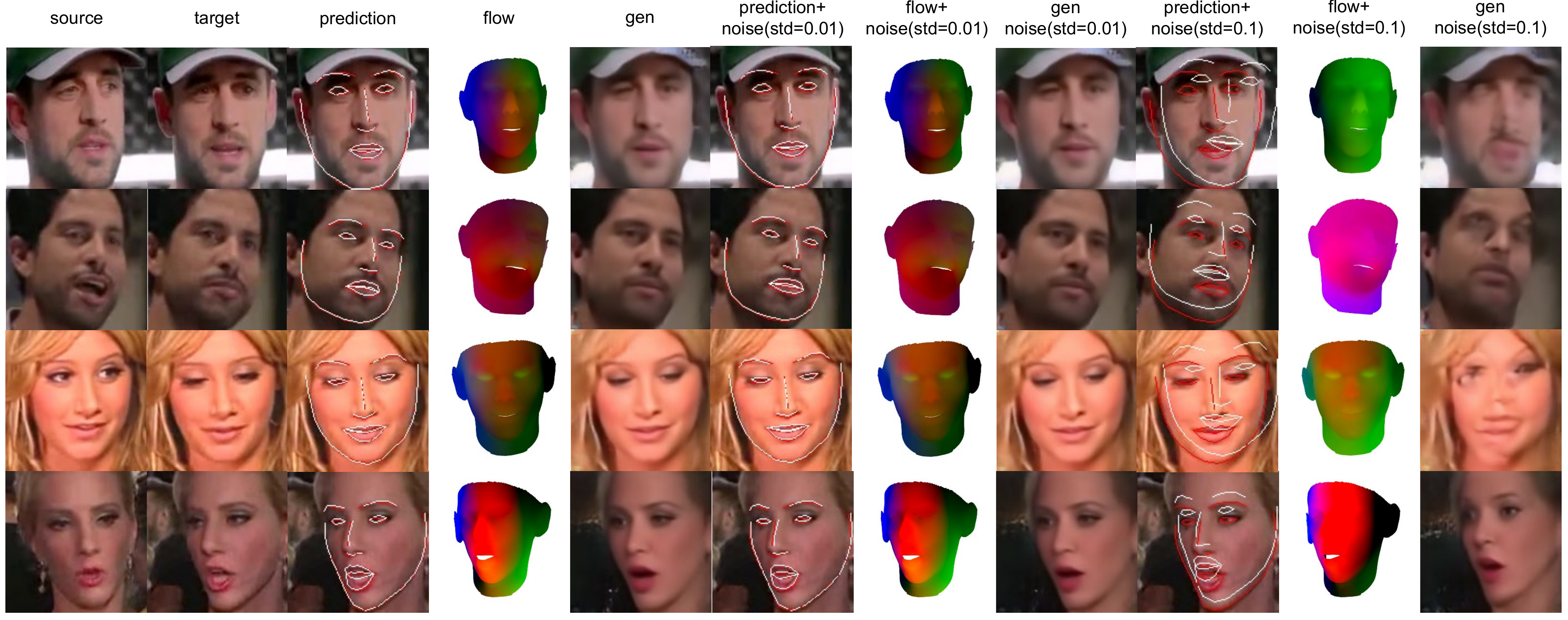}
		\caption{Qualitative results of facial flows with different degrees of noise on Voxceleb2 dataset. The 1st and 2nd column refers to the source and target ground-truth image. The 3rd, 6th, 9th column refers to the prediction of 3D face landmarks with noise. Red: ground truth from [66]. White: predictions of the 3DMM regressor model. The 4th, 7th, 10th column refers to the flow with noise. The 5th, 8th, 11th column refers to the generated images with noise.}
		\label{fig:noise_flows}
	\end{figure*}
	
	\subsection{Cross-Reenactment}
	
	Another interesting aspect in face manipulation is the ability to transfer the pose or expression of other identity to the source one, named cross-reenactment. In fact, it is difficult to get the ground-truth images of the source identity with target expression from the driven identity. Previous works have studied this as a cycle-consistency problem \cite{pumarola2020ganimation, zhu2017unpaired}, training a cyclegan with unpaired data. However, this will increase a lot of extra calculations in the training process because two generators are needed in this setting. Our method only needs a forward generator, which was trained with paired data of the same identity. Although the generator has not seen the cross-id data when training, we can also easily achieve the cross-id reenactment due to the disentanglement of 3DMM parameters. As introduced in \cref{III-A}, 3DMM parameters $P$ consists of shape parameters $\alpha_{id}$, expression parameters $\alpha_{exp}$, pose parameters $r$ and other transformation parameters. Given source and target images of different identities, when $\alpha_{id}$ is fixed, we can substitute $\alpha_{exp}$ and $r$ selectively from the driven image to generate a new facial flow for the source identity. Thus, a source face with generated facial flow can be sent into $\textup{F}^{\textup{3}}$A-GAN for cross identity reenactment. 
	
	\mbox{\cref{fig:vox2_reenact}}-(b) and \mbox{\cref{tab:voxceleb2}}-(right part) show the qualitative and quantitative results of cross-reenactment on Voxceleb2 dataset, respectively. Whether in self-reenactment or cross-reenactment, our method is able to transform the pose and expression well. In \mbox{\cref{tab:voxceleb2}}-(right part), X2Face keeps the identity well but fails to keep the pose since it cannot adapt to large pose transformation well. Pix2pixHD is able to transform the pose and expression well but fail to maintain the identity. Our method achieves the trade-off between pose transformation and identity preservation.

	Moreover, our method is able to reenact the input image in a controllable manner, \textit{i.e.}, performing expression or pose reenactment only. Please see the qualitative results in Section II in supplemental material.


	\subsection{Ablation Study}
	
	Our method consists of two main components, Facial Flow Constructor and Hierarchical Conditional Framework. We make ablation study on both of them including the comparison on condition selection and architectural variants.
	
	\subsubsection{Conditions Study} Prior models such as action units and landmark are widely used in face animation. We compare our proposed facial flow with these two conditions under our framework. To unify the input format, we spatially replicate the action units vector (17-d) and then channel-wisely concatenate it with the replicated Euler angle vector (3-d) to form an input tensor ($H\times W\times 20$). For landmarks, we follow \cite{songsri2019face} and transform it to a spatial heatmap ($H\times W\times 1$).
	
	Compared results can be seen in \cref{fig:conditions}. As it shows in the third row, conditional frameworks based on action units and Euler angles fail to generate the right face with the left face as input. This might be the reason that the 3-d vector is too sparse to model the whole pose transformation. For the others, conditions based on geometric prior including landmark heatmaps and our proposed facial flow adapt to the large-pose transformation well under our proposed framework. However, as highlighted in the red box, we observe that weird eyes (shorter eye spacing) are generated in the second row of the first example. This is because that linear interpolation in the 2D plane will produce landmarks that do not conform to the real face movement, which will be solved in facial flow. Also, it could be seen from the second example that our proposed condition beats the others.
	
	Moreover, we calculated the quantitative results on different conditions in \mbox{\cref{tab:self-300VW}}. The results show that the proposed facial flows outperform action units and landmarks in each metrics since we model the dense geometric information in facial flows for better representation. The proposed facial flows are able to better maintain the pose and identity.
	
	\begin{table}[!t]
		\centering
		\caption{Quantitative results of facial flows with different degrees of noise on Voxceleb2 dataset. It evaluates the robustness of the whole system based on facial flows.}
		\begin{tabular}{c|ccccc}
			\toprule
			\multicolumn{1}{c}{} & MAE$\downarrow$ & LMAE$\downarrow$ & PSNR$\uparrow$ & SSIM$\uparrow$ & CSIM$\uparrow$ \\
			\midrule
			Ours  & \textbf{0.0571} & \textbf{0.0066} & \textbf{21.7062} & \textbf{0.7426} & \textbf{0.9082} \\
			\midrule
			noise (std=0.01) & 0.0584 & 0.0092 & 21.5099 & 0.7333 & 0.9056 \\
			\midrule
			noise (std=0.1) & 0.0880 & 0.0554 & 18.1745 & 0.5521 & 0.7988 \\
			\bottomrule
		\end{tabular}%
		\label{tab:noise_flows}%
	\end{table}%
	
	\subsubsection{Architectural Variants} There exists some other architectural variants in our proposed framework. We compare the quality results in \cref{fig:variants} and analyze them one by one in the following.
	
	\textbf{One-stream manipulation $T_{os}$}. We adopt the concatenation between input images and conditions directly and present a one-stream transformer. In $T_{os}$, the convolutional filters process the information from images and facial flows at the same time, which tend to learn the average of these two instead of their independent information. This will increase the training difficulties and lead to artifacts. However, in our method, a two-stream network structure can better extract the dynamic and static features simultaneously.
	
	\textbf{Single layer manipulation $T_{sl}$}. In order to verify the effectiveness of the hierarchical manipulation strategy, we present a compared manipulation framework that only manipulates the features in the latent space. As the results show, single layer manipulation is also capable of manipulating the source face to the target one but fail to preserve the details in ears and hair. This is because semantic features in high-level layers lack detailed information. 
	
	\textbf{Non-progressive manipulation $T_{np}$}. We also train the network without progressive manipulation. The network output the final image of $128\times 128$ size directly. As shown in the fourth column in \cref{fig:variants}, the framework without progressive strategy outputs images of accurate expression and pose but lower quality. 
	
	Moreover, we calculated the quantitative results on different model architectures in \mbox{\cref{tab:self-300VW}}. The lower MAE and MSE metrics illustrate that the two-stream and progressive training strategies of our method enables the model to generate high-quality images.
	
	\begin{figure}[!t]
		\centering
		\includegraphics[width=\linewidth]{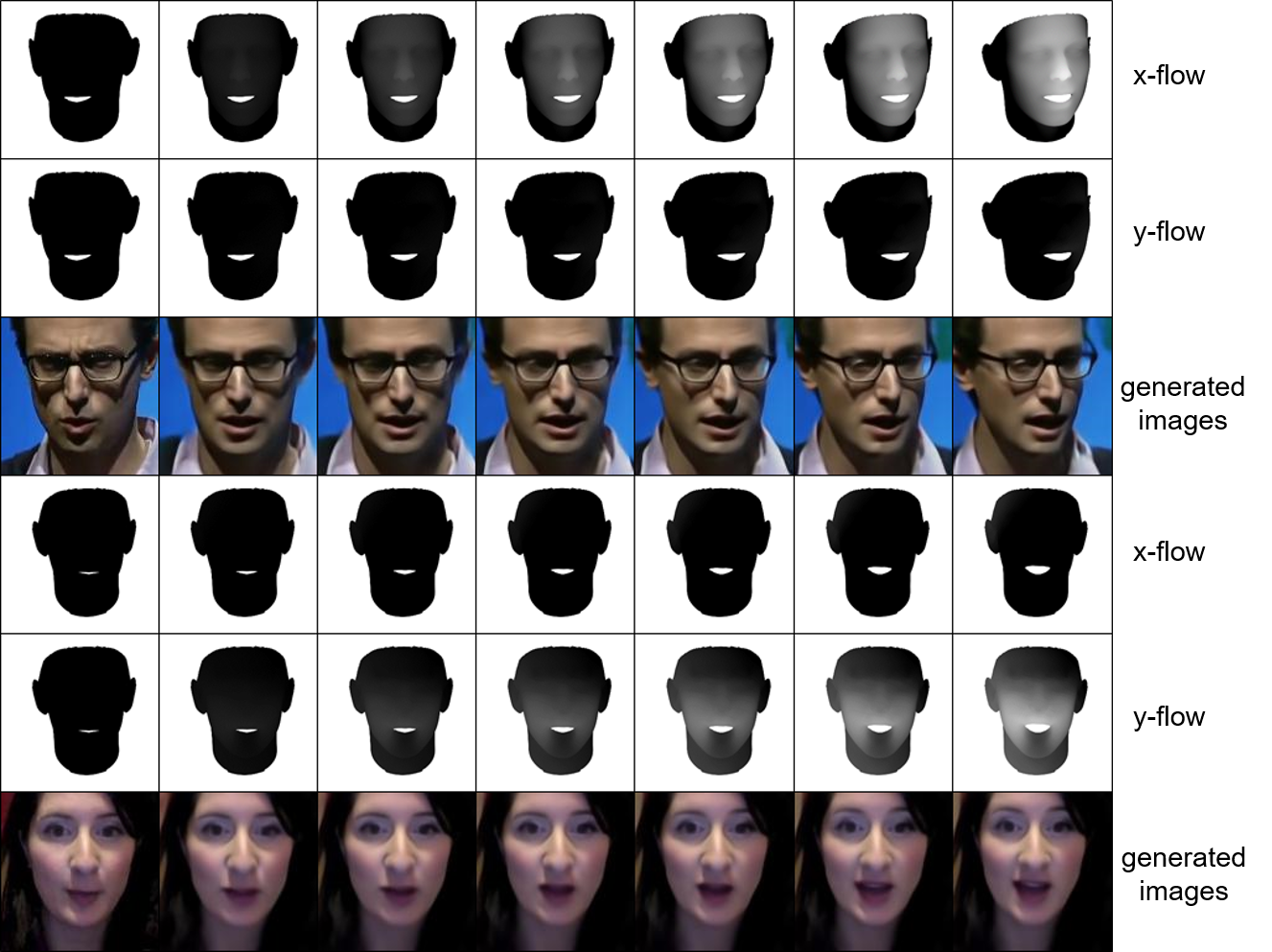}
		\caption{Visualization results of the generated output images with facial flow components along different axis. We named the facial flow components along the $x,y,z$ axis in the three-dimensional space x-flow, y-flow, z-flow, respectively. Every three rows correspond to a sample. The first row shows the changes of x-flow while the second row is for y-flow. The third row shows our synthesized images generated by the conditional framework taking facial flows as conditions.}
		\label{fig:flowxy}
	\end{figure}
	
	\subsection{Analyzing Facial Flow}
	
	In this subsection, we analyze the smoothness and continuity of the proposed facial flow qualitatively, and estimate its impact on the whole system. Firstly, since the expression and pose changes are finally reflected in the 2D image, we visualize the flow map in $x,y$ axis. As shown in the first example in \cref{fig:flowxy}, the man is performing his head's movement from left to right, so his most motion happens in $x$ axis. We can observe the obvious bright pattern in x-flow picture indicating large motion in this axis. Contrarily, the woman in the second sample keeps her head fixed so little motion can be seen from the x-flow picture. While she is opening her mouth, we can observe a bright pattern in her lips area in y-flow picture. Upon her lips is darker than the mean color while below shallower which indicated the separation of two lips. Our facial flow can represent the motion information in high fidelity and guarantee the synthesized results with great geometrical accuracy.
	
	Secondly, we conduct noise analysis on the 3DMM parameter regressor model. Specifically, we perform different degrees of noise \textit{i.e.}, gaussian noise with std=0.01 and std=0.1, on the parameters extracted from 3DMM regressor model and show the corresponding results of the whole system in \mbox{\cref{tab:noise_flows}} and \mbox{\cref{fig:noise_flows}}. As shown in \mbox{\cref{fig:noise_flows}}, when the extracted facial flow describes the expression and pose of the target image accurately, the generated image is consistent with the target one (See the 3rd-5th columns). When adding noise on the 3DMM parameters, the expression and pose information in facial flow will be distorted, and it will cause the system to generate images with inconsistent goals. The impact of adding different degrees of noise is listed in \mbox{\cref{tab:noise_flows}}. The generated images deviate gradually away from the ground truth as added noise increases.

	\section{Conclusion}
	In this paper, we propose a two-stage framework --- $\textup{F}^{\textup{3}}$A-GAN for face animation. $\textup{F}^{\textup{3}}$A-GAN consists of two main components: 1) facial flow constructor for estimating a facial flow between a source and a target face image; 2) hierarchical conditional framework combining the facial flow and source image for target image generation. The facial flow integrates two superior factors, \textit{i.e.,} natural motion and geometric continuity into a spatial map. The hierarchical conditional framework combines the appearance features in the image stream and motion features in the flow stream hierarchically and outputs the images progressively. With the linearity and disentanglement of our facial flow, the framework is capable of synthesizing continuous images well and controlling the pose or expression reenactment. Experimental results show that our method outperforms other synthesis methods in terms of both the visual quality and the diversity of generation. 
	
	\section*{Acknowledgement}
	The authors would like to thank Bin Li for his valuable suggestions.

	
	%

	%
	%
	%
	%
	%

	\ifCLASSOPTIONcaptionsoff
	\newpage
	\fi

	\normalem
	\balance
	\bibliographystyle{ieeetr}
	\bibliography{IEEEabrv,reference}
	
	%
	%
	%
	
	
	

\end{document}


%
\title{Supplementary Material:\\
	\huge{``$\textup{F}^{\textup{3}}$A-GAN: Facial Flow for Face Animation with Generative Adversarial Networks"}}
\author{
	Xintian Wu, Qihang Zhang, Yiming Wu, Huanyu Wang, Songyuan Li, Lingyun Sun and Xi~Li
}

\maketitle
\renewcommand{\theequation}{S-\arabic{equation}}
\setcounter{equation}{0}
\renewcommand{\thefigure}{S-\arabic{figure}}
\setcounter{figure}{0}
\renewcommand{\thetable}{S-\arabic{table}}
\setcounter{table}{0}

\section{Implementaion Details} \label{implement_detail}
The network training in our method is a two-stage learning scheme. In this section, we will describe the architectural and training details of the networks in the following. 

\subsection{3DMM Regressor} We train a CNN-based 3DMM regressor to obtain the 3DMM parameters of input images and then fit the 3D face models with these regressed parameters. We extend the encoder in PRNet [54] with two fully connected layers to predict 3DMM parameters because we need to generate various facial flows through manipulating the 3DMM parameters.

We train such a CNN regressor in a semi-supervised manner with labeled image dataset (300W-3D) and unlabeled video datasets (300VW / voxceleb2). To specifically implement the training process, we train the network for 20 epochs, using Adam optimizer with a learning rate of 0.0001, beta1 0.9, beta2 0.999 and batch size 256. In the loss $L_{lm}$, we increased the weight mask by 10 times in the mouth area and 5 times in the eyes area. The super parameters $\lambda_1, \lambda_2, \lambda_3$ are set to be 10, 1, 1e-3.

\subsection{Conditional Framework} The hierarchical conditional framework consists of an encoder $T_{enc}$ and decoder $T_{dec}$ in the image stream, and a flow driver $T_{flow}$ in the flow stream. The $T_{enc}$ and $T_{dec}$ are combined as an auto-encoder manner. We mainly made the following modifications to the traditional auto-encoder structure. Firstly, we replace the pooling layers with stride convolutional layers in $T_{enc}$ and utilize stride deconvolutional layers in $T_{dec}$ for feature resolution doubling because the pooling layers will cause CNNs to lose translation invariance. Secondly, the encoder structure in $T_{flow}$ is similar to that in $T_{enc}$ but with fewer channels. Thirdly, in order to ensure the independence of a single image transformation, all the Batch Normalization layers are replaced by the Instance Normalization layers because the latter only performs normalization operations on a single image. Last, we set PReLU as the activation layers to improve training stability.

Both the generator and the discriminator were trained progressively with 40 epochs. For other settings, we used an Adam optimizer with learning rate 1e-4, beta1 0.5, beta2 0.999, and batch size was set as 64. In order to allow the discriminator to be more fully trained for metric learning, every 2 optimization steps of the discriminator we performed a single optimization step of the generator. The weight coefficients for $L_{tot}$ were set as $\lambda_4 = 100$, and $\lambda_5 = 10$.

\vspace{2em}
\section{Qualitative Results on Generation} \label{animation}
\begin{figure}[htbp]
	\centering
	\includegraphics[width=\linewidth]{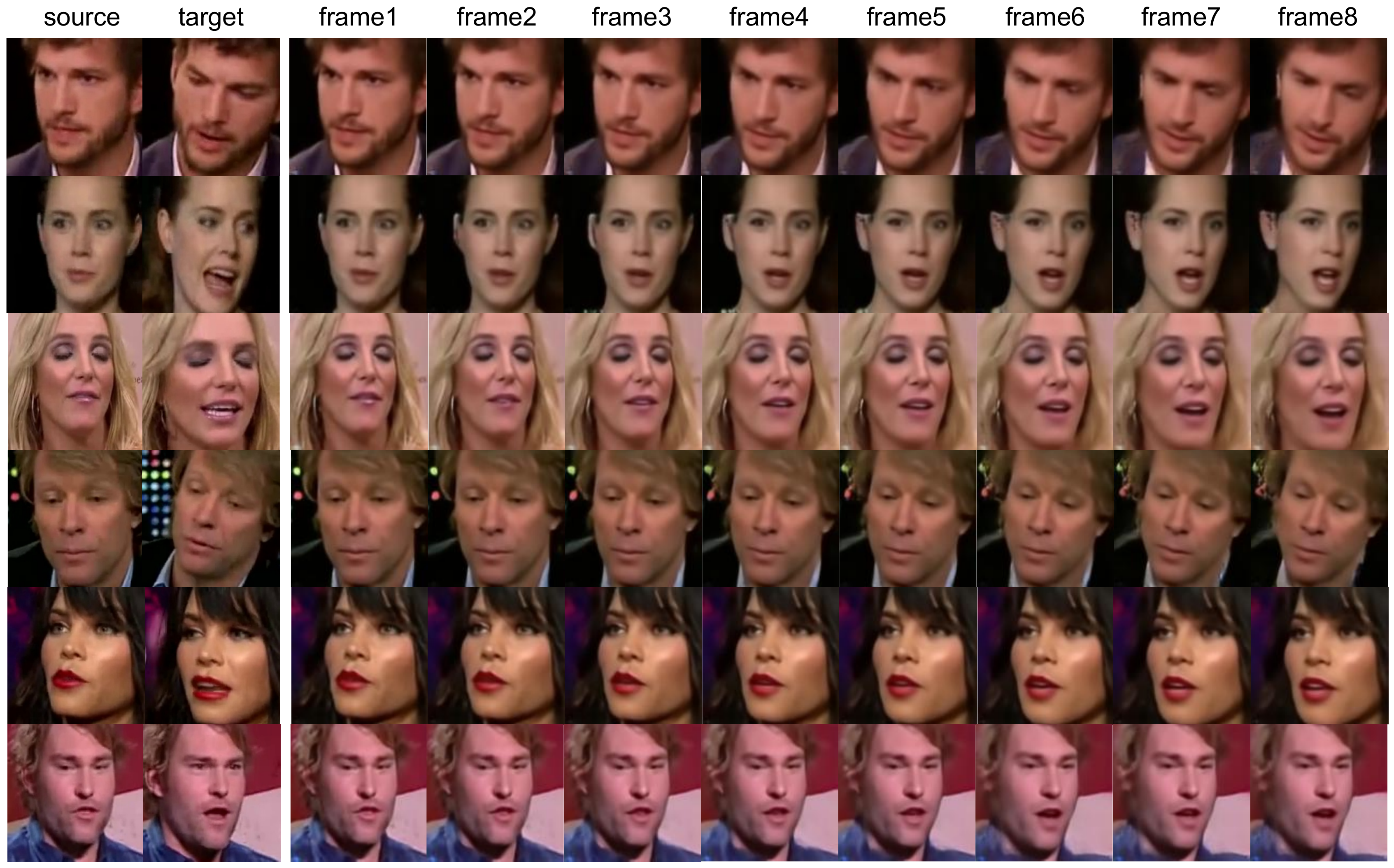}
	\caption{Qualitive results of our method on continuous face generation. In these examples, we do interpolation on the corresponding conditions between two images $I_s$ (first column) and $I_t$ (second column) to generate the intermediate continuous images (from third column to the last column).}
	\label{fig:face_animation_vox2}
\end{figure}

\begin{figure}[htbp]
	\centering
	\includegraphics[width=\linewidth]{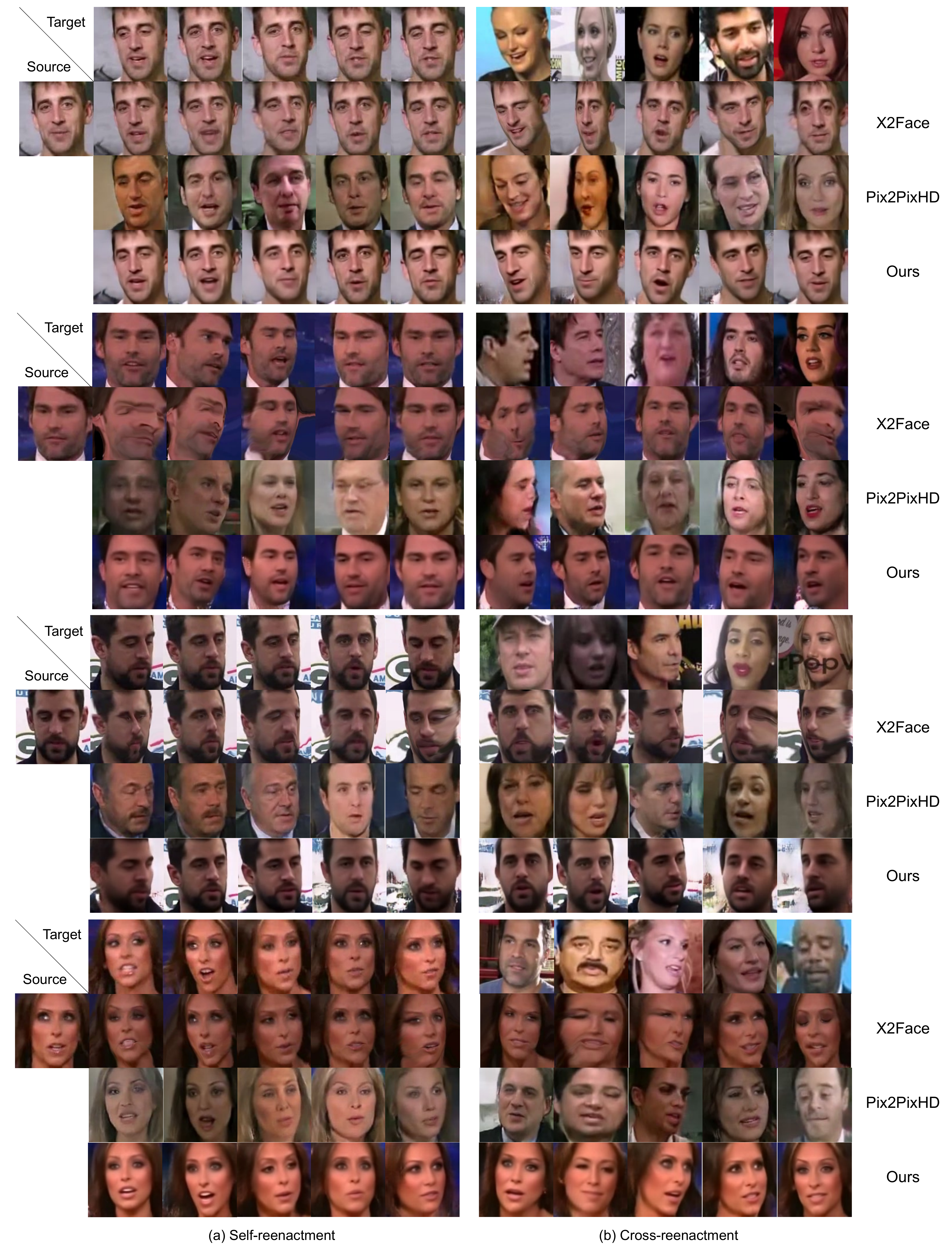}
	\caption{Qualitative results of self-reenactment and cross-reenactment on Voxceleb2 dataset. For self (a) / cross (b) identity reenactment, we randomly select few target images with the same / different identity as source image to drive the model generation. Each row indicates the generated images of the same method.}
	\label{fig:vox2_reenact}
\end{figure}

In this section, we show more results in three test scenarios, continuous image generation , self and cross identity reenactment and controllable reenactment. \cref{fig:face_animation_vox2} illustrates more examples of continuous generation through our method. The results show that our method is able to anime the source image for generating continuous images smoothly. \cref{fig:vox2_reenact} illustrates more examples of self and cross identity reenactment through our method. 

\begin{figure}[htbp]
	\centering
	\includegraphics[width=\linewidth]{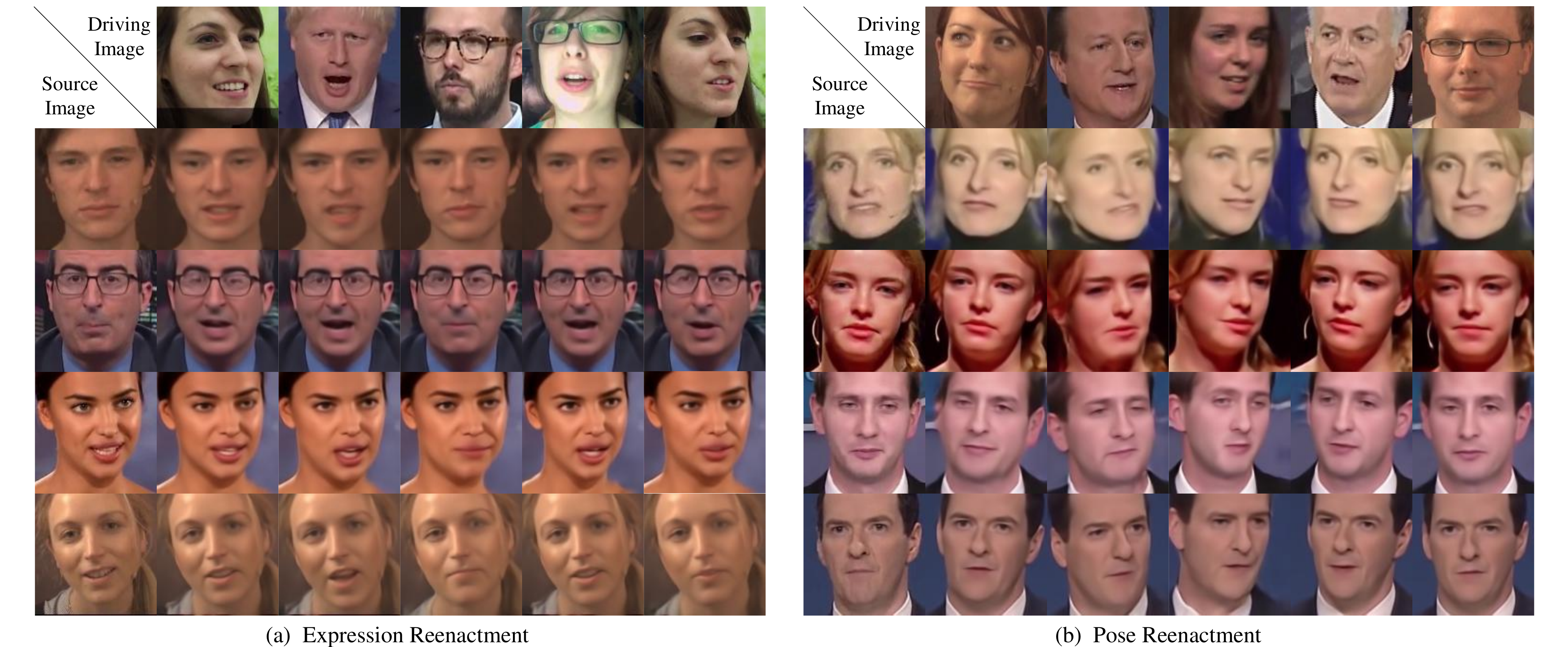}
	\caption{Qualitative results of our proposed methods in controllable face reenactment on 300VW dataset. (a) illustrates expression reenactment for various source image (first column) and driving image (first row). (b) illustrates the pose reenactment.}
	\label{fig:face_reenact_300VW}
\end{figure}

Moreover, we also report the qualitative results in controllable reenactment in \cref{fig:face_reenact_300VW}. It  demonstrates multiple manipulation outputs controlled by expression or pose only. The results reveal the controllable ability of our method. For instance, in expression reenactment, although the driven faces perform different poses from the source one, they only transfer the expression successfully while maintaining the pose well, and vice versa. 

\newpage
\section{Good and Bad Cases}

\begin{figure}[htbp]
	\centering
	\includegraphics[width=0.4\linewidth]{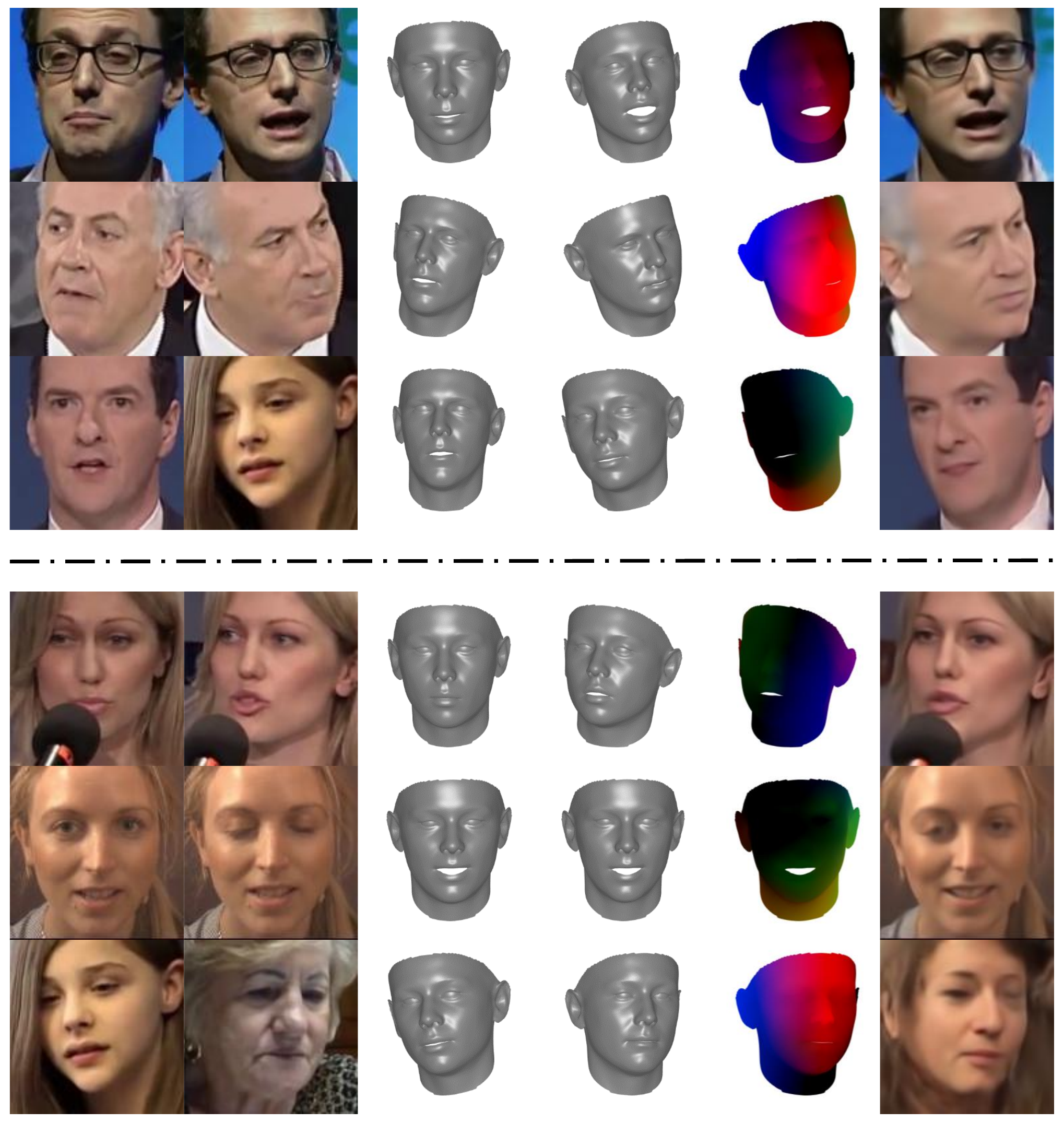}
	\caption{Good and bad cases. Cases on top of the dotted line are good cases while bellow are bad cases. In all cases, from left to right, we represent the source image $I_s$, the target image $I_t$, the reconstructed source 3D face model $X^s$, the target model $X^t$, the facial flow $f$ and the generated output $T(I_s, f)$.}
	\label{fig:fail}
\end{figure}

Finally, we push the limits of our framework and discuss the model limitations. We demonstrate both the good cases (top-rows) and bad cases (bottom-rows) in \cref{fig:fail}. In good cases, the generator is able to output the translation results with driven expression and pose well. For the first row in \cref{fig:fail}-top, although the 3D face model does not model the glasses, the generator can imagine the correct position of the glasses. For the second row, even in the case large-pose transformation, the output result seems to be transformed correctly. For the third row, it can also transfer the expression and pose from different identity. In bad cases, we analyze the limitations of our framework. Since all the 3D scans for 3DMM are with eyes open, it is difficult for a 3DMM to fit a face with eyes closed. As shown in the first and second row in \cref{fig:fail}-bottom, the generator cannot generate images with eyeball moving and eyes closing. In addition, there still exists an overfitting problem that in some cases, the method cannot reenact well. For example, a poor output is demonstrated in the third row. In fact, the girl in the video hardly rotates her head and this leads to many redundant training pairs. We will try to solve this problem by introducing few-shot learning in future study.

\ifCLASSOPTIONcaptionsoff
  \newpage
\fi